\newif\ifjournal\journalfalse           % techreport style

\ifjournal
\documentclass[numreferences]{kluwer}
\usepackage{graphicx}
\usepackage{amsmath}
\usepackage{amssymb}
\else
\documentclass[12pt,twoside]{article}
\usepackage{latexsym}
\usepackage{graphicx}
\usepackage{amsmath,amssymb}
\usepackage{amsthm}
\makeindex
\fi

\edef\,{\thinspace}\edef\;{\thickspace}\edef\!{\negthinspace}
\def\dispmuskip{\thinmuskip= 3mu plus 0mu minus 2mu \medmuskip=  4mu plus 2mu minus 2mu \thickmuskip=5mu plus 5mu minus 2mu}
\def\textmuskip{\thinmuskip= 0mu                    \medmuskip=  1mu plus 1mu minus 1mu \thickmuskip=2mu plus 3mu minus 1mu}
\textmuskip
\def\beq{\dispmuskip\begin{eqnarray}}    \def\eeq{\end{eqnarray}\textmuskip}
\def\beqn{\dispmuskip\begin{displaymath}}\def\eeqn{\end{displaymath}\textmuskip}
\def\bqa{\dispmuskip\begin{eqnarray}}    \def\eqa{\end{eqnarray}\textmuskip}
\def\bqan{\dispmuskip\begin{eqnarray*}}  \def\eqan{\end{eqnarray*}\textmuskip}

\newtheorem{theorem}{Theorem}

\newtheorem{proposition}[theorem]{Proposition}
\newtheorem{definition}[theorem]{Definition}

\ifjournal
\def\qedx{\hspace*{\fill}$\Box\quad$}
\newproof{proof}{Proof}
\else

\def\qedx{}
\newenvironment{keywords}%
  {\centerline{\bf\small Keywords}\begin{quote}\small}%
  {\par\end{quote}\vskip 1ex}
\fi

\ifjournal\def\paragraph#1{\noindent\subparagraph{#1.}}
\else\def\paragraph#1{\vspace{1.3ex plus 1 ex minus 1 ex}\noindent{\bf{#1.}}}
\fi

\def\odt{{\textstyle{1\over 2}}}

\def\v{\boldsymbol} %\boldsymbol\frak\Bbb\pmb\text
\def\p{{\scriptscriptstyle+}}
\def\pp{{\scriptscriptstyle++}}

\def\u{u}
\def\kdot{{\dot\kappa}}
\def\vt{{\v t}}\def\vu{{\v\u}}\def\vpi{{\v\pi}}

\def\Var{{\mbox{Var}}}
\def\Cov{{\mbox{Cov}}}
\def\qmbox#1{{\quad\mbox{#1}\quad}}
\def\low{\underline}
\def\up{\overline}
\def\argmin{\mathop{\rm arg\,min}}          % minarg
\def\undertilde#1{{\smash{\mathop{#1}\limits_{\widetilde{}}}}{}}
\def\underbartilde#1{{\smash{\mathop{#1}\limits_\simeq}}{}} %undertilde and underbar
\def\upa{\overline}
\def\lowa{\underline}

\begin{document}
%%%%%%%%%%%%%%%%%%%%%%%%%%%%%%%%%%%%%%%%%%%%%%%%%%%%%%%%%%%%%%%%%
%                      T i t l e - P a g e                      %
%%%%%%%%%%%%%%%%%%%%%%%%%%%%%%%%%%%%%%%%%%%%%%%%%%%%%%%%%%%%%%%%%

\ifjournal
\begin{article}
\begin{opening}
\title{Robust Inference of Trees}
\author{Marco \surname{Zaffalon}\email{zaffalon@idsia.ch}\thanks{Research partially supported by the Swiss NSF grant 2100-067961.}\\
        Marcus \surname{Hutter}\email{marcus@idsia.ch}\thanks{Research partially supported by the Swiss NSF grant 2000-61847.00 to J\"{u}rgen Schmidhuber.}}
\institute{IDSIA, Galleria 2, CH-6928\ Manno (Lugano), Switzerland}

\else

\title{\vskip -3mm\bf\Large\hrule height5pt \vskip 5mm
\sc Robust Inference of Trees
\vskip 5mm \hrule height2pt \vskip 5mm}
\author{{\bf Marcus Hutter}\thanks{Research partially supported by the Swiss NSF grant 2100-067961.}
\ and {\bf Marco Zaffalon}\thanks{Research partially supported by the Swiss NSF grant 2000-61847.00 to J\"{u}rgen Schmidhuber.}\\[3mm]
\normalsize IDSIA, Galleria 2, CH-6928\ Manno-Lugano, Switzerland\\
\normalsize \{zaffalon,marcus\}@idsia.ch \hspace{7ex} http://www.idsia.ch }
\date{\normalsize Submitted: August 2002 \hspace{2ex} In Press: December 2005}
\maketitle
\fi

\begin{abstract}
This paper is concerned with the reliable inference of optimal
tree-approximations to the dependency structure of an unknown
distribution generating data. The traditional approach to the
problem measures the dependency strength between random variables
by the index called \emph{mutual information}. In this paper reliability
is achieved by Walley's \emph{imprecise Dirichlet model}, which
generalizes Bayesian learning with Dirichlet priors. Adopting the
imprecise Dirichlet model results in posterior interval
expectation for mutual information, and in a set of plausible
trees consistent with the data. Reliable inference about the
actual tree is achieved by focusing on the substructure common to
all the plausible trees. We develop an exact algorithm that infers
the substructure in time $O(m^4)$, $m$ being the number of random
variables.  The new algorithm is applied to a set of data sampled
from a known distribution. The method is shown to reliably infer
edges of the actual tree even when the data are very scarce,
unlike the traditional approach. Finally, we provide lower and
upper credibility limits for mutual information under the
imprecise Dirichlet model. These enable the previous developments
to be extended to a full inferential method for trees.
\end{abstract}

\ifjournal
\keywords{Robust inference, spanning trees, intervals,
dependence, graphical models, mutual information, imprecise
probabilities, imprecise Dirichlet model.}
\classification{AMS classification}{05C05, 41A58, 62G35, 62H20,
68T37, 90C35.}
\end{opening}
\else
\begin{keywords}
Robust inference, trees, spanning trees, intervals, dependence,
graphical models, mutual information, imprecise probabilities,
imprecise Dirichlet model.
\end{keywords}
\pagebreak{\parskip=0ex\tableofcontents}
\fi

%%%%%%%%%%%%%%%%%%%%%%%%%%%%%%%%%%%%%%%%%%%%%%%%%%%%%%%%%%%%%%%
\section{Introduction}\label{secInt}
%%%%%%%%%%%%%%%%%%%%%%%%%%%%%%%%%%%%%%%%%%%%%%%%%%%%%%%%%%%%%%%

This paper deals with the following problem. We are given a random
sample of $n$ observations, which are jointly categorized
according to a set of $m$ nominal random variables $\imath$,
$\jmath$, $\kappa$, etc. The dependency between two variables is
measured by the information-theoretic symmetric index called
\emph{mutual information} \cite{Kul68}. If the chances\footnote{We
denote vectors by $\v x:=(x_1,...,x_d)$ for $\v x\in\{\v
n,\vt,\vu,\vpi,...\}$.} $\vpi$ of all instances defined by the
co-occurrence of $\imath=i$, $\jmath=j$, $\kappa=\kdot$, etc.,
were known, it would be possible to approximate the distribution
by another, for which all the dependencies are bivariate and can
graphically be represented as an undirected tree $T$, that is the
optimal approximating tree-dependency distribution
(Section~\ref{secB}). This result is due to Chow and Liu
\cite{ChowLiu68}, who use Kullback-Leiber's divergence
\cite{KulLei51} to measure the similarity of two distributions.

Since only a sample is available, the joint distribution $\vpi$ is
unknown and an inferential approach is necessary. Prior
uncertainty about the vector $\vpi$ is described by the
\emph{imprecise Dirichlet model} (IDM) \cite{Wal96b}. This is an
inferential model that generalizes Bayesian learning with
Dirichlet priors, by using a set of prior densities to model prior
(near-)ignorance. Using the IDM results in posterior uncertainty
about $\vpi$, the mutual information and the tree $T$
(Section~\ref{secB}). In general, this makes a set of trees
$\mathcal{T}$ consistent with the data.

Robust inference about $T$ is achieved by identifying the edges
common to all the trees in $\mathcal{T}$, called \emph{strong
edges} (Sections~\ref{secSBWG}).  An exact and an approximate
algorithm are developed that detect strong edges in times
$O(m^4)$ and $O(m^3)$, respectively. The
former is applied to a set of data sampled from a known
distribution, and is compared with the original algorithm from
Chow and Liu (Section~\ref{secAE}). The new algorithm is shown to
reliably infer partial trees (we call them \emph{forests}), which
quickly converge to the actual complete tree as the sample grows.
Unlike the traditional approach based on precise probabilities,
the new algorithm avoids drawing wrong edges by suspending the
judgement on those for which the information is poor.

Many technical issues are addressed in the paper to develop the
new algorithm. The identification of strong edges involves solving
a problem on graphs. We develop original exact and approximate
algorithms for this task in Section~\ref{secSBWG}. Robust
inference involves computing bounds for the lower and upper
expectation of mutual information under the IDM
(Section~\ref{secRCOE}). We provide conservative (i.e.,
over-cautious) bounds that at most make an error of magnitude
$O(n^{-2})$.

These results lead to important extensions, reported in
Section~\ref{secE}. Inference on mutual information is extended by
providing lower and upper credibility limits under the IDM (i.e.,
intervals that depend on a given guarantee level). The overall
approach extends accordingly. Furthermore, we discuss alternatives
to the strong edges algorithm proposed in this paper, aiming to
exploit the results presented here in wider contexts.

To our knowledge, the literature only reports two other attempts
to infer robust structures of dependence. Kleiter \cite{Kleiter99}
uses approximate confidence intervals on mutual
information\footnote{Note that accurate expressions for credible
mutual information intervals have been derived in
\cite{Hut01,HutZaf05}.} to measure the dependence between random
variables. Kleiter's work is different in spirit from ours. We
look for tree structures that are optimal in some sense, by using
systematic and reliable interval approximations to the actual
mutual information. Kleiter focuses on general graphical
structures and is not concerned with questions of optimality.
Bernard \cite{Ber01} describes a method to
build a directed graph from a multivariate binary database. The
method is based on the IDM and Bayesian implicative analysis. The
connection with our work is looser here since the arcs of the
graph are interpreted as logical implications rather than
probabilistic dependencies.

%%%%%%%%%%%%%%%%%%%%%%%%%%%%%%%%%%%%%%%%%%%%%%%%%%%%%%%%%%%%%%%
\section{Background}\label{secB}
%%%%%%%%%%%%%%%%%%%%%%%%%%%%%%%%%%%%%%%%%%%%%%%%%%%%%%%%%%%%%%%

%-------------------------------------------------------------%
\subsection{Maximum spanning trees}\label{MST}
%-------------------------------------------------------------%

This paper is concerned with trees. In the undirected case, trees
are undirected connected graphs with $m$ nodes and $m-1$ edges.
Undirected trees are such that for each pair of nodes there is
only one path that connects them \cite[Proposition~2]{PapSte82}.
Directed trees can be constructed from undirected ones, orienting
the arrows in such a way that each node has at most a single
direct predecessor (or \emph{parent}). When used to represent
dependency structures, the nodes of a tree are regarded as random
variables and the tree itself represents the dependencies between
the variables. It is a well-known result that all the directed
trees that share the same undirected structure represent the same
set of dependencies \cite{VerPea90}. This is the reason why the
inference of directed trees from data focuses on recovering the
undirected structure; and it is also the reason why this paper is
almost entirely concerned with undirected trees (called more
simply `trees' in the following).

Chow and Liu \cite{ChowLiu68} address the problem of approximating
the actual pattern of dependencies of a distribution by an
undirected tree. Their work is based on mutual information. Given
two random variables $\imath$, $\jmath$ with values in
$\{1,...,d_\imath\}$ and $\{1,...,d_\jmath\}$, respectively, the
\emph{mutual information} is defined as
\beqn {\cal I}({\vpi}) \;=\; \sum_{i=1}^{d_\imath}\sum_{j=1}^{d_\jmath}
  \pi_{ij}\log{\pi_{ij}\over\pi_{i\p}\pi_{\p j}} \;,
\eeqn where $ \pi_{ij} $ is the actual chance of $ (i,j) $, and
$\pi_{i\p}:=\sum_j\pi_{ij}$ and $\pi_{\p j}:=\sum_i\pi_{ij}$ are
marginal chances. Chow and Liu's algorithm works by computing the
mutual information for all the pairs of random variables. These
values are used as edge \emph{weights} in a fully connected graph.
The output of the algorithm is a tree for which the sum of the
edge weights is maximum. In the literature of graph algorithms,
the general version of the last problem is called the
\emph{maximum spanning tree} \cite[p.~271]{PapSte82}. Its
construction takes $O(m^{2})$ time. This is also the computational
complexity of the above procedure. The tree constructed as above
is shown to be an optimal tree-approximation to the actual
dependencies when the similarity of two distributions is measured
by Kullback-Leiber's divergence \cite{KulLei51}.

Chow and Liu extend their procedure to the inference of trees from
data by replacing the mutual information with the \emph{sample
mutual information} (or \emph{empirical mutual information}). This
approximates the actual mutual information by using, in the
expression for mutual information, the sample relative frequencies
instead of the chances $\pi_{ij}$, which are typically unknown in
practice.

%-------------------------------------------------------------%
\subsection{Robust inference}\label{sectRI}
%-------------------------------------------------------------%

Using empirical approximations for unknown quantities, as
described in the previous section, can lead to fragile models.
Fragile models produce quite different outputs depending on the
random fluctuations involved in the generation of the sample.

Reliability can be achieved by robust inferential tools. In this
paper we consider the imprecise Dirichlet model
\cite{Wal96b,bernard2005}. The IDM is a model of inference for
multivariate categorical data. It models prior uncertainty using a
set of Dirichlet prior densities and does posterior inference by
combining them with the likelihood function (see
Section~\ref{secIDM} for details). The IDM rests on very weak
prior assumptions and is therefore a very robust inferential tool.

The IDM leads to lower and upper expectations for mutual
information (and, possibly, lower and upper credibility limits),
i.e., to intervals. This is a complication for the discovery of
tree structures from data. In fact, the maximum spanning tree
problem assumes that the edge weights can be totally ordered. Now,
multiple values of mutual information are generally consistent
with the given intervals. In general, this prevents us from having
a total order on the edges: not all the pairs of edges can be
compared.

The generalization of Chow and Liu's approach is achieved via the
definition of more general graphs that can deal with multiple edge
weights. This is done in the next section.

%%%%%%%%%%%%%%%%%%%%%%%%%%%%%%%%%%%%%%%%%%%%%%%%%%%%%%%%%%%%%%%
\section{Set-based weighted graphs}\label{secSBWG}
%%%%%%%%%%%%%%%%%%%%%%%%%%%%%%%%%%%%%%%%%%%%%%%%%%%%%%%%%%%%%%%

Consider an undirected fully connected graph $G_w=<V,E>$, with
$m=\left|V\right|$ nodes, and where $E$ denotes the set of edges
[$(v,v)\notin E$ for each $v\in V$]. $G_w$ is also a weighted
graph, in the sense that each edge $e\in E$ is associated with the
real number $w(e)$, which in this paper will be a value of mutual
information. Consider a set of graphs with the same topological
structure but different weight functions $w$ in a non-empty set
$W$: $\mathcal{G}=\{G_w: w\in W\}$. We call $\mathcal{G}$ a
\emph{set-based weighted graph}. Note that $\mathcal{G}$ can be
thought of also as a single graph $G$, on each edge $e$ of which
there is a set of real weights: $\{w(e): w\in W\}$. Yet, for the
latter view to be equivalent to the former, one should pay
attention to the fact that there could be logical dependencies
between weights of two different sets; in other words, it could be
the case that not all the pairs of weights in the cartesian
product of two sets appear in a single graph of $\mathcal{G}$.

In order to extend the notion of maximum spanning tree to
set-based weighted graphs, we define the solution of the maximum
spanning tree problem generalized to set-based weighted graphs, as
the set $\mathcal{T}$ of maximum spanning trees originated by the
graphs in $\mathcal{G}$.

Recall that Kruskal's algorithm only needs a total order on the
edges to build a unique maximum spanning tree \cite{Kruskal56}.
Therefore, in order to focus on $\mathcal{T}$, we can equivalently
focus on the set $\mathcal{O_T}$ of total orders that are
consistent with the graphs in $\mathcal{G}$. In the following we
find it more convenient not to directly deal with $\mathcal{O_T}$.
Rather, we first show how to construct a partial order that is
consistent with all the total orders in $\mathcal{O_T}$, and then
we consider all the total orders that extend the partial order.
Initially, we need the following definition.

\begin{definition}
We say that edge $e$ \emph{dominates} edge $e^{\prime }$ if
$w(e)>w(e^{\prime})$ for all $w\in W$.\label{dominance}
\end{definition}

By applying the above definition to all the distinct pairs of
edges in $G$ we obtain the sought partial order. To see that the
order is only partial in general, consider the example graph in
Figure~\ref{fig10}. We have defined such a graph $G$ by drawing
the graphical structure and specifying set-based weights by
placing intervals on the edges in a separate way (i.e., assuming
logical independency between different intervals). That is, the
example graph is equivalent to the set $\mathcal{G}$ of graphs
obtained by choosing real weights within the intervals in all the
possible ways. Now observe that the intervals for the edges (A,B)
and (B,C) overlap, so that there is no dominance in either
direction. Figure~\ref{fig20} shows the overall partial order on
the edges for the graph in Figure~\ref{fig10}.

\begin{figure}[h]\begin{center}
  \includegraphics[width=4cm]{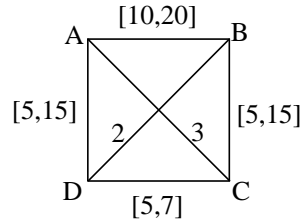}\\
  \caption{An example set-based weighted graph. The sets for the
edges are specified separately by intervals that in two cases
degenerate to real numbers.}\label{fig10}
\end{center}\end{figure}

\begin{figure}[h]\begin{center}
  \includegraphics[width=8cm]{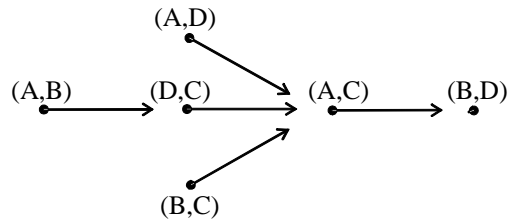}\\
  \caption{The partial order on the edges of the graph in the
preceding figure. Here an arrow from $e$ to $e^{\prime }$ means
that $e$ dominates $e^{\prime }$.}\label{fig20}
\end{center}\end{figure}

Now we consider the set $\mathcal{O}$ of all the total orders that
extend the partial order induced by Definition~\ref{dominance}. Of
course, $\mathcal{O}$ includes $\mathcal{O_T}$. They coincide if
for each total order in $\mathcal{O}$, there is a graph
$G_w\in\mathcal{G}$ in which $w(e)>w(e^{\prime})$ if $e$
dominates $e^{\prime}$ in the given total order. This is the case,
for example, when mutual information is separately specified via
intervals on the edges.

%-------------------------------------------------------------%
\subsection{Exact detection of strong edges}\label{secEDOSE}
%-------------------------------------------------------------%

We call \emph{strong edges} the edges of $G$ that are common to
all the trees in $\mathcal{T}$. Identifying the strong edges
allows us to robustly infer dependencies that belong to the
unknown optimal approximating trees. The following theorem is the
central tool for the identification.

\begin{theorem}
\label{th1}Assume $\mathcal{O} =\mathcal{O_T}$. An edge $e$ of $G$ is
strong if and only if in each simple\footnote{This is a cycle in
which the nodes are all different. In the following we will simply
refer to simple cycles as cycles.} cycle that contains $e$ there
is an edge $e^{\prime}$ dominated by $e$.
\end{theorem}

\begin{proof}
\end{proof}

($\Leftarrow $) By contradiction, assume that there is a graph
$G_w\in\mathcal{G}$ for which an optimal tree $T$ does not contain
$e$. By adding $e$ to $T$ we create a cycle
\cite[Proposition~2]{PapSte82}. By hypothesis, in such a cycle
there must exist an edge $e^{\prime}$ dominated by $e$, so
$w(e)>w(e^{\prime})$. Removing $e^{\prime}$, we obtain a new tree
that improves upon $T$, so that $T$ cannot be optimal for $G_w$.

($\Rightarrow $) By contradiction, assume that there is a cycle
$C$ in $G$ where $e$ does not dominate any edge. Then there is a
total order in $\mathcal{O}$ in which $e$ is dominated by any
other edge $e^{\prime}$ in $C$. Since $\mathcal{O}
=\mathcal{O_T}$, there must also exist a related graph $G_w$ for
which $w(e)\le w(e^{\prime})$ for any edge $e^{\prime}$ in $C$.
Call $T$ the related tree. By removing $e$ from $T$ we create two
subtrees, say $T^{\prime }$ and $T^{\prime \prime }$. One of these
can possibly be a degenerate tree composed by a single node. Now
consider that there must be an edge $e_{C}$ of $C$ that connects a
node of $T^{\prime}$ with one of $T^{\prime\prime}$. If there was
not, there would be no way to start from an endpoint of $e$ in
$T^{\prime}$ and reach the other endpoint, because all the paths
would be confined within $T^{\prime}$. The graph composed by
$T^{\prime }$, $T^{\prime \prime }$ and $e_{C}$ has $m-1$ edges,
spans all the nodes of $G$, and therefore it is a tree, say $T^*$
\cite[Proposition~2]{PapSte82}. If $w(e)<w(e_{C})$, $T^*$
improves upon $T$, so that $T$ cannot be optimal for $G_w$. If
$w(e)=w(e_{C})$, both $T^*$ and $T$ are optimal, but their
intersection does not contain $e$, so $e\notin\mathcal{T}$.
%\qedx\end{proof}

Theorem~\ref{th1} directly leads to a procedure that determines
whether or not a given edge $e$ is strong. It suffices to consider
the graph $G'$ obtained from $G$ by removing $e$ and the edges
that $e$ dominates (see the Procedure `DetectStrongEdges' in Table
\ref{strong edges}). Edge $e$ is strong if and only if its
endpoints are not connected in $G'$. By applying this procedure to
the graph in Figure~\ref{fig10}, we conclude that only (A,B) is
strong.

Note that Theorem~\ref{th1} assumes that $\mathcal{O}$ coincides
with $\mathcal{O_T}$. If this failed to be true,
$\mathcal{O_T}\subset\mathcal{O}$ would still hold, making
Theorem~\ref{th1} work with a set of trees larger than
$\mathcal{T}$, eventually leading to an excess of caution: the
edges determined by the above procedure would anyway be strong,
but there might be strong edges that the procedure would not be
able to determine.

As for computational considerations, note that testing whether or
not two nodes are connected in a graph demands $O(m^{2})$ time.
Repeating the test for all the edges $e\in E$, we have the
computational complexity of the overall procedure, $O(m^{4})$.

%-------------------------------------------------------------%
\subsection{Approximate detection of strong edges}\label{secADOSE}
%-------------------------------------------------------------%

This section presents a procedure that approximately detects the
strong edges, reducing the complexity to $O(m^{3})$ with respect
to the exact procedure given in Section~\ref{secEDOSE}.

Consider the algorithm outlined in a pseudo programming language
in Table~\ref{alg1}. It takes as input a fully connected graph
$G=<V,E>$. In the algorithm, a tree with a number of nodes in
$\{2,\ldots ,m-1\}$ is called \emph{subtree}.

\begin{table}
\begin{enumerate}
\item[ ]

\item  \label{start}Let $SE=\emptyset $;

\item  \label{first loop}for each $v\in V$

\begin{enumerate}
\item  \label{first test}if there is a node $v^{\prime }\in V$ such that
$(v,v^{\prime })\notin SE$ and it dominates $(v,v^{\prime \prime })$ for each
$v^{\prime\prime }\in V$, $v^{\prime \prime }\neq v^{\prime }$ then

\begin{enumerate}
\item  \label{first insertion}add $(v,v^{\prime })$ to $SE$;
\end{enumerate}

\end{enumerate}

\item  if there is a subtree in $SE$ then

\begin{enumerate}
\item  \label{make it}make it the current subtree;

\item  \label{E primo}consider the set of edges $E^{\prime}\subseteq E$
with one endpoint in the nodes of the current subtree and the other outside;

\item  \label{second test}if there is an edge $e^{\prime}\in E^{\prime }$
that dominates all the other edges in  $E^{\prime }$
then

\begin{enumerate}
\item  \label{second insertion}add $e^{\prime }$ to $SE$ and to the current subtree;

\item  \label{first go to} go to~\ref{E primo};
\end{enumerate}

\item  else

\begin{enumerate}
\item  if there is another subtree in $SE$ not considered yet then

\begin{enumerate}
\item  \label{second go to} go to~\ref{make it};
\end{enumerate}

\item  else output $SE$.

\end{enumerate}

\end{enumerate}

\end{enumerate}
\caption{Approximate procedure to detect strong edges.}\label{alg1}
\end{table}

The following proposition shows that the algorithm in
Table~\ref{alg1} returns only strong edges.

\begin{proposition}
\label{prop0}$SE$ is a subset of the strong edges of $G$.
\end{proposition}

\begin{proof}

Consider the first possible insertion in Step~\ref{first
insertion}. The cycles that encompass $(v,v^{\prime })$ must pass
through the set of edges $\{(v,v^{\prime \prime }):v^{\prime
\prime }\in V,v^{\prime \prime }\neq v^{\prime }\}$. Since
$(v,v^{\prime })$ dominates all the edges in the preceding set,
for each cycle passing through $(v,v^{\prime })$ there is an edge
in the cycle that is dominated by $(v,v^{\prime })$, so that
$(v,v^{\prime })$ is strong, by Theorem~\ref{th1}.

The algorithm can insert an edge in $SE$ also in Step~\ref{second
insertion}. Recall that each subtree is a connected acyclic graph.
It is clear that any cycle that contains $e^{\prime }$ must pass
through an edge $e^{\prime \prime }$ that has one endpoint in the
nodes of the subtree and the other outside. But $e^{\prime }$
dominates $e^{\prime \prime }$ by Step~\ref {second test}. This
holds for all the cycles, so $e^{\prime }$ is strong by Theorem~\ref{th1}.
\qedx\end{proof}

The logic of the algorithm in Table~\ref{alg1} is to move from
subtrees made of strong edges to adjacent nodes, in order to
detect the strong edges of a graph. This policy does not allow all
the strong edges to be determined in general. For example, the
approximate algorithm cannot determine that the edge (A,B) in
Figure~\ref{fig10} is strong.

The heuristic policy implements a trade-off between computational
complexity and the capability to fully detect the strong edges.
This choice does not seem critical to the specific extent of
discovering tree-dependency structures. In fact, the knowledge of
the actual mutual information increases with the sample size,
becoming a number in the limit. It is easy to check that in these
conditions the exact and the approximate procedure produce the
same set of edges.

%-------------------------------%
\paragraph{Computational complexity}\label{CC}
%-------------------------------%
The assumption behind the following analysis is that the
comparison of two edges can be done in constant time. In this
case, given a set $E^{\prime }$ of edges, there is a procedure
that determines in time $O(\left| E^{\prime }\right| )$ if there
is an edge $e^{\prime }\in E^{\prime }$ that dominates all the
others. The first step of the procedure selects an edge that is
candidate to be dominant. This is made by doing pairwise
comparisons of edges and by always discarding the non-dominant
edge (or edges) in the comparison. After at most $\left|
E^{\prime}\right| -1$ comparisons, we know whether there is a
candidate or not. If there is, the second step of the procedure
compares such candidate $e^{\prime }$ with all the other edges,
deciding if $e^{\prime }$ dominates all the others. This requires
$\left| E^{\prime }\right| -1$ comparisons. The two steps of the
procedure take $O(\left| E^{\prime }\right| $) time.

Let us now focus on the algorithm in Table~\ref{alg1}. The
loop~\ref{first loop} is repeated $m=\left| V\right| $ times. Each
time the test~\ref{first test} decides whether there is a dominant
edge out of $m-1$ edges (each node is connected to all the
others). By the previous result, such task takes $O(m)$ time. Then
the loop requires $O(m^{2})$ time.

Now consider the two nested loops made by the
instructions~\ref{make it}, \ref{E primo}, \ref{first go to},
and~\ref{second go to}. Each time the instruction of
jump~\ref{first go to} is executed, a new edge has been added to
$SE$. Each time~\ref{second go to} is executed, a new subtree is
considered. Since $SE$ can have $m-1$ edges at most and $m$ is
also an upper bound on the number of different subtrees, the two
loops can jointly require $2m-1$ iterations at most. Each such
iteration executes the test~\ref{second test}. By using $m^{2}$ as
an upper bound on $\left| E^{\prime }\right| $, we need $O(m^{2})$
time to detect whether the dominant edge exists. The overall time
required by the loops is $O(m^{3})$. This is also the
computational complexity of the entire procedure.

%%%%%%%%%%%%%%%%%%%%%%%%%%%%%%%%%%%%%%%%%%%%%%%%%%%%%%%%%%%%%%%
\section{Robust comparison of edges}\label{secRCOE}
%%%%%%%%%%%%%%%%%%%%%%%%%%%%%%%%%%%%%%%%%%%%%%%%%%%%%%%%%%%%%%%

So far we have focused on the detection of strong edges, taking
for granted that there exists a method to partially compare edges
based on imprecise knowledge of mutual information. We provide
such a method in the following sections. We will first present a
formal introduction to the imprecise Dirichlet model in
Section~\ref{secIDM}. Section~\ref{secREE} will make a first step
by computing robust estimates for the entropy. These will be used
in Section~\ref{secREMI} to derive robust estimates of mutual
information. Finally, the method to compare edges will be given in
Section~\ref{secCE}.

%-------------------------------------------------------------%
\subsection{The imprecise Dirichlet model}\label{secIDM}
%-------------------------------------------------------------%

%-------------------------------%
\paragraph{Random i.i.d.\ processes}
%-------------------------------%
We consider a discrete random variable $\imath$ and
a related i.i.d.\ random process with samples $i\in\{1,...,d\}$ drawn
with probability $\pi_i$. The chances $\vpi$ form a probability
distribution, i.e.,\ $\vpi\in\Delta:=\{\v x\in I\!\!R^d\,:\,x_i\geq
0\,\forall i,\; x_\p=1\}$, where we have used the abbreviation
$x_\p:=\sum_{i=1}^d x_i$. The likelihood of a specific data set
$\v D=(i_1,...,i_n)$ with $n_i$ samples $i$ and total sample size
$n=n_\p=\sum_i n_i$ is $p(\v D|\vpi)\propto\prod_i\pi_i^{n_i}$.
Quantities of interest are, for instance, the entropy ${\cal
H}(\vpi)=-\sum_i\pi_i\log\pi_i$, where $\log$ denotes the natural
logarithm. The chances $\pi_i$ are usually unknown and have to be
estimated from the data.

%-------------------------------%
\paragraph{Second order p(oste)rior}
%-------------------------------%
In the Bayesian approach one models the initial uncertainty in
$\vpi$ by a (second order) prior distribution $p(\vpi)$ with
domain $\vpi\in\Delta$. The Dirichlet priors
$p(\vpi)\propto\prod_i\pi_i^{n'_i-1}$, where $n'_i$ comprises
prior information, represent a large class of priors. $n'_i$ may
be interpreted as (possibly fractional) ``virtual'' sample
numbers. High prior belief in $i$ can be modelled by large $n'_i$.
It is convenient to write $n'_i=s\cdot t_i$ with $s:=n'_+$, hence
$\vt\in\Delta$. Examples for $s$ are $0$ for Haldane's prior
\cite{Haldane:48}, $1$ for Perks' prior \cite{Perks:47}, ${d\over
2}$ for Jeffreys' prior \cite{Jeffreys:46}, and $d$ for
Bayes-Laplace's uniform prior \cite{Gelman:95} (all with
$t_i={1\over d}$). These are also called \emph{noninformative
priors}. From the prior and the data likelihood one can determine
the posterior $p(\vpi|\v D)=p(\vpi|\v
n)\propto\prod_i\pi_i^{n_i+st_i-1}$. The expected value or mean
$\u_i:=E_\vt[\pi_i]={n_i+st_i\over n+s}$ is often used for
estimating $\pi_i$ (the accuracy may be obtained from the
covariance of $\vpi$). The expected entropy is $E_\vt[{\cal
H}]=\int_\Delta {\cal H}(\vpi)p(\vpi|\v n)d\vpi$. An approximate
solution can be obtained by exchanging $E$ with ${\cal H}$ (exact
only for linear functions): $E_\vt[{\cal H}(\vpi)]\approx {\cal
H}(E_\vt[\vpi])={\cal H}(\vu)$. The approximation error is
typically of the order ${1\over n}$. In
\cite{Wolpert:95,Hut01,HutZaf05} exact expressions have been
obtained:
\bqa\label{hex}
  E_\vt[{\cal H}] &=& H(\vu) \;:=\; \sum_i h(\u_i) \qmbox{with}
  \\\nonumber
  h(\u) &=& \u\!\cdot\![\psi(n+s+1)-\psi((n+s)\u+1)],
\eqa
where $\psi(x)=d\,\log\Gamma(x)/dx$ is the logarithmic derivative of
the Gamma function. There are fast implementations of $\psi$ and
its derivatives and exact expressions for integer and
half-integer arguments (see Appendix~\ref{secPsi}).

%-------------------------------%
\paragraph{Definition of the imprecise Dirichlet model}
%-------------------------------%
There are several problems with noninformative priors. First, the
inference generally depends on the arbitrary definition of the
sample space. Second, they assume exact prior knowledge $p(\vpi)$.
The solution to the second problem is to model our ignorance by
considering sets of priors $p(\vpi)$, a model that is part of the
wider theory of \emph{imprecise}\footnote{In the following we will
avoid the term {\em imprecise} in favor of {\em robust}, since
expressions like ``exact imprecise intervals'' sound confusing.}
\emph{probabilities} \cite{Wal91}. The specific imprecise
Dirichlet model \cite{Wal96b} considers the set of
all\footnote{Strictly speaking, $\Delta$ should be the open
simplex \cite{Wal96b}, since $p(\vpi)$ is improper for $\vt$ on
the boundary of $\Delta$. For simplicity we assume that, if
necessary, considered functions of $\vt$ can be, and are, continuously
extended to the boundary of $\Delta$, so that, for instance,
minima and maxima exist. All considerations can straightforwardly,
but cumbersomely, be rewritten in terms of an open simplex. Note
that open/closed $\Delta$ result in open/closed robust intervals,
the difference being numerically/practically irrelevant.}
$\vt\in\Delta$, i.e., $\{p(\vpi):\vt\in\Delta\}$, which solves also
the first problem. Walley suggests to fix the hyperparameter $s$
somewhere in the interval $[1,2]$. A set of priors results in a
set of posteriors, set of expected values, etc. For real-valued
quantities like the expected entropy $E_\vt[{\cal H}]$ the sets
are typically intervals:
\beqn
  E_\vt[{\cal H}] \;\in\; [\min_{\vt\in\Delta}E_\vt[{\cal H}] \,,\,
    \max_{\vt\in\Delta}E_\vt[{\cal H}]] \;=:\;
   [\low H,\up H].
\eeqn
In the next section we derive approximations for
\beqn
   \up H \;=\; \max_{\vt\in\Delta} H(\vu) \qmbox{and}
  \low H \;=\; \min_{\vt\in\Delta} H(\vu).
\eeqn
One can show that $h(\u)$ is strictly concave (see
Appendix~\ref{secPsi}), i.e., \ $h''(\u)<0$ and that $h''$ is
monotone increasing ($h'''>0$), which we exploit in the following.
The results for the entropy serve as building blocks to derive
similar results for the needed mutual information.
We define the general correspondence
\beqn
 \u_i^{\cdots}={n_i+st_i^{\cdots}\over n+s},
 \quad\mbox{where $^{\ldots}$ can be various superscripts}.
\eeqn

%-------------------------------------------------------------%
\subsection{Robust entropy estimates}\label{secREE}
%-------------------------------------------------------------%

%-------------------------------%
\paragraph{Taylor expansion of $H(\vu)$}
%-------------------------------%
In the following we derive reliable approximations for $\up H$ and
$\low H$. If $n$ is not too small these approximations are close
to the exact values. More precisely, the length of interval $[\low
H,\up H]$ is $O(\sigma)$, where $\sigma:={s\over n+s}$, while the
approximations will differ from $\up H$ and $\low H$ by at most
$O(\sigma^2)$. Let $t_i^*\in[0,1]$ and $\u_i^*={n_i+st_i^*\over
n+s}$. This implies
\beq\label{dtbnd}
  \u_i-\u_i^* \;=\; \sigma\!\cdot\!(t_i-t_i^*) \qmbox{and}
  |\u_i-\u_i^*| \;=\; \sigma|t_i-t_i^*| \;\leq\; \sigma.
\eeq
Hence we may Taylor-expand $H(\vu)$ around $\vu^*$. $H$ is
approximately linear in $\vu$ and hence in $\vt$. A linear
function on a simplex assumes its extreme values at the vertices
of the simplex.
The most natural point for expansion is $t_i^*={1\over d}$ in the
center of $\Delta$. For this choice the bound~(\ref{dtbnd}) and
most of the following bounds can be improved to
$\sigma\leadsto\sigma|1-{1\over d}|$. Other, even data-dependent
choices like $t_i^*={n_i\over n}=\u_i^*$, are possible. The only
property we use in the following is that\footnote{$\arg\min_i n_i$
is the $i$ for which $n_i$ is minimal. Ties can be broken
arbitrarily. Kronecker's $\delta_{i,j}=1$ for $i=j$ and
$\delta_{i,j}=0$ for $i\neq j$.} $\arg\max_i\u_i^*=\arg\max_i
n_i$ and $\arg\min_i\u_i^*=\arg\min_i n_i$. We have
\beqn
  H(\vu) \;=\; \overbrace{H(\vu^*)}^{H_0=O(1)} +
  \overbrace{\sum_i h'(\u_i^*)(\u_i-\u_i^*)}^{H_1=O(\sigma)} +
  \overbrace{\odt\sum_i
  h''(\check\u_i)(\u_i-\u_i^*)^2}^{H_R=O(\sigma^2)}.
\eeqn
For suitable $\check\u_i$ between $\u_i^*$ and $\u_i$ this
expansion is exact ($H_R$ is the exact remainder).

%-------------------------------%
\paragraph{Approximation of $\up H$}
%-------------------------------%
Inserting~(\ref{dtbnd}) into $H_1$ we get
\beqn
  H_1 \;=\; \sum_i h'(\u_i^*)(\u_i-\u_i^*) \;=\;
  \sigma\sum_i h'(\u_i^*)(t_i-t_i^*).
\eeqn
Ignoring the $O(\sigma^2)$ remainder $H_R$, in order to maximize
$H(\vu)$ we only have to maximize $\sum_i h'(\u_i^*)t_i$ (the only
$\vt$-dependent part). A linear function on $\Delta$ is maximized
by setting the $t_i$ component with largest coefficient to 1. Due
to concavity of $h$, $h'(\u_i^*)$ is largest for the smallest
$\u_i^*$, i.e., \ for smallest $n_i$, i.e., \ for $i=\upa
i:=\arg\min_i n_i$. Hence $\up{H_1}=H_1(\upa\vu)$, where $\upa
t_i:=\delta_{i,\upa i}$ and $\upa\vu$ follows from $\upa\vt$ by the
general correspondence. $H_0+\up{H_1}$ is an $O(\sigma^2)$
approximation of $\up H$. Consider now the remainder $H_R$:
\beqn
  H_R \;=\; \odt\sigma^2\sum_i h''(\check\u_i)|t_i-t_i^*|^2
      \;\leq\; 0 \;=:\; H_R^{ub}
\eeqn
due to $h''<0$. This bound cannot be improved in general, since
$H_R=0$ is attained for $t_i=t_i^*$. Non-positivity of $H_R$ shows
that $H_0+\up{H_1}$ is an upper bound of $\up H$. Since
$\up H\geq H(\vu)$ for all $\vu$, $H(\upa\vu)$ in
particular is a lower bound on $\up H$, and moreover also an
$O(\sigma^2)$ approximation. Together we have
\beqn
  \underbrace{H(\upa\vu)}_{\up H-O(\sigma^2)} \;\leq\;
  \up H \;\leq\; \underbrace{H_0 + \up{H_1}}_{\up H+O(\sigma^2)}.
\eeqn
For robust estimates, the upper bound is, of course, more
interesting.

%-------------------------------%
\paragraph{Approximation of $\low H$}
%-------------------------------%
The determination of $\low{H_1}$ follows the same scheme
as for $\up{H_1}$. We get $\low{H_1}=H_1(\lowa\vu)$ with
$\lowa t_i:=\delta_{i,\lowa i}$ and $\lowa i:=\arg\max_i n_i$.
Using $|t_i-t_i^*|\leq
1$, $\check\u_i\geq{n_i\over n+s}$, $h''<0$ and that $h''$ is
monotone increasing ($h'''>0$) we get the following lower bound on
the remainder $H_R$:
\beqn
  H_R \;=\; \odt\sigma^2\sum_i h''(\check\u_i)|t_i-t_i^*|^2
  \;\geq\; \odt\sigma^2\sum_i h''(\textstyle{n_i\over n+s}) \;=:\; H_R^{lb}.
\eeqn
Putting everything together we have
\beqn
  \underbrace{H_0 + \low{H_1}}_{\low H-O(\sigma^2)} +
  \underbrace{H_R^{lb}}_{O(\sigma^2)}
  \;\leq\; \low H \;\leq\;
  \underbrace{H(\lowa\vu)}_{\low H+O(\sigma^2)}.
\eeqn
For robust estimates, the lower bound is more interesting. General
approximation techniques for other quantities of interest are
developed in \cite{Hut03}. Exact expressions for $[\low H,\up H]$
are also derived there.

%-------------------------------------------------------------%
\subsection{Robust estimates for mutual information}\label{secREMI}
%-------------------------------------------------------------%

%-------------------------------%
\paragraph{Mutual information}
%-------------------------------%
Here we generalize the bounds for the entropy found in
Section~\ref{secREE} to the mutual information of two random
variables $\imath$ and $\jmath$ that take values in
$\{1,...,d_\imath\}$ and $\{1,...,d_\jmath\}$, respectively.
Consider an i.i.d.\ random process with samples
$(i,j)\in\{1,...,d_\imath\}\times\{1,...,d_\jmath\}$ drawn with
joint probability $\pi_{ij}$, where $\vpi\in\Delta :=\{\v x\in
I\!\!R^{d_\imath\times d_\jmath}\,:\,x_{ij}\geq 0\,\forall ij,\;
x_\pp=1\}$. We are interested in the mutual information of
$\imath$ and $\jmath$:
\bqan
  {\cal I}({\vpi}) &=& \sum_{i=1}^{d_\imath}\sum_{j=1}^{d_\jmath}
  \pi_{ij}\log{\pi_{ij}\over\pi_{i\p}\pi_{\p j}}
\\\nonumber &=&
  \sum_{ij}\pi_{ij}\log\pi_{ij} -
  \sum_{i}\pi_{i\p}\log\pi_{i\p} -
  \sum_{j}\pi_{\p j}\log\pi_{\p j}
\\\nonumber &=&
  {\cal H}(\vpi_{\imath\p}) +
  {\cal H}(\vpi_{\p\jmath}) -
  {\cal H}(\vpi_{\imath\jmath}).
\eqan
$\pi_{i\p}=\sum_j\pi_{ij}$ and $\pi_{\p j}=\sum_i\pi_{ij}$ are
marginal probabilities. Again, we assume a Dirichlet prior over
$\vpi_{\imath\jmath}$, which leads to a Dirichlet posterior
$p(\vpi_{\imath\jmath}|\v
n)\propto\prod_{ij}\pi_{ij}^{n_{ij}+st_{ij}-1}$. The expected value
of $\pi_{ij}$ is
\beqn
 \u_{ij}:=E_\vt[\pi_{ij}]={n_{ij}+st_{ij}\over n+s}.
\eeqn
The marginals $\vpi_{i\p}$ and $\vpi_{\p j}$ are also
Dirichlet
with expectation $\u_{i\p}$ and $\u_{\p j}$.
The expected mutual information $E_\vt[{\cal I}]$
can, hence, be expressed in terms of the expectations of
three entropies
\beqn
  I(\vu) := H(\vu_{\imath\p})+H(\vu_{\p\jmath})-H(\vu_{\imath\jmath})
 \;=\; H_{left}+H_{right}-H_{joint}
\eeqn
\beqn
 \;=\; \sum_i h(\u_{i\p})+\sum_j h(\u_{\p j})-\sum_{ij}h(\u_{ij})
\eeqn
where here and in the following we index quantities with {\em
joint}, {\em left}, and {\em right} to denote to which
distribution the quantity refers. Using~(\ref{hex}) we get
$E_\vt[{\cal I}]=I(\vu)$.

%-------------------------------%
\paragraph{Crude bounds for $I(\vu)$}
%-------------------------------%
Estimates for the IDM interval $[\min_{\vt\in\Delta}E_\vt[{\cal
I}]$, $\max_{\vt\in\Delta }E_\vt[{\cal I}]]$ can be obtained by
minimizing/maximizing $I(\vu)$. A crude upper bound can be
obtained as
\beqn
  \up I \;:=\; \max_{\vt\in\Delta} I(\vu) \;=\;
  \max[H_{left}+H_{right}-H_{joint}] \;\leq\;
\eeqn
\beqn
  \max H_{left} + \max H_{right} - \min H_{joint} \;=\;
  \up H_{left}  + \up H_{right}  - \low H_{joint},
\eeqn
where upper and lower bounds to $\up H_{left}$, $\up H_{right}$ and
$\low H_{joint}$ have been derived in Section~\ref{secREE}.
Similarly
$
  \low I \geq \low H_{left} + \low H_{right} - \up H_{joint}
$.
The problem with these bounds is that, although good in some
cases, they can become arbitrarily crude. In the following we
derive bounds similar to the entropy case with $O(\sigma^2)$
accuracy.

%-------------------------------%
\paragraph{$O(\sigma^2)$ bounds for $I(\vu)$}
%-------------------------------%
We expand $I(\vu)$ around $\vu^*$ with a constant term
$I_0$, a term $I_1$ linear in $\sigma$ and an exact $O(\sigma^2)$
remainder.
\beqn
  I(\vu) \;=\; I_0+I_1+I_R, \qquad
  I_0=H_{0left}+H_{0right}-H_{0joint} = I(\vu^*),
\eeqn
\bqan
  & I_1 & =\; H_{1left}+H_{1right}-H_{1joint}
\\[1ex]
  &=& \sum_i h'(\u_{i\p}^*)(\u_{i\p}\!\!-\!\u_{i\p}^*) +
  \sum_j h'(\u_{\p j}^*)(\u_{\p j}\!-\!\u_{\p j}^*) -
  \sum_{ij} h'(\u_{ij}^*)(\u_{ij}\!-\!\u_{ij}^*)
\\
  &=& \sigma\sum_{ij} g_{ij}(t_{ij}-t_{ij}^*)
  \qmbox{with}
  g_{ij}:=h'(\u_{i\p}^*)+h'(\u_{\p j}^*)-h'(\u_{ij}^*).
\eqan
$I_1$ is maximal if $\sum_{ij} g_{ij}t_{ij}$ is maximal. This is
maximal if $t_{ij}=\upa t_{ij}:=\delta_{(ij),\upa{(ij)}}$ and
$\upa{(ij)}:=\arg\max_{(ij)} g_{ij}$, hence $\up{I_1}=I_1(\upa\vu)$, and
$I_0+\up{I_1}$ and $I(\upa\vu)$ being $O(\sigma^2)$ approximations
to $\up I$. Replacing all max's by min's we get $I_0+\low{I_1}$
and $I(\lowa\vu)$ as $O(\sigma^2)$ approximations to $\low I$. To get
robust bounds we need bounds on $I_R =
H_{R\,left}+H_{R\,right}-H_{R\,joint}. $
\bqan
  I_R & \leq &
  \max_{\vu,\v{\check\u}}[H_{R\,left}+H_{R\,right}-H_{R\,joint}]
  \\ & \leq &
  H_{R\,left}^{ub}+H_{R\,right}^{ub}-H_{R\,joint}^{lb}
  \;=\; -H_{R\,joint}^{lb} =: I_R^{ub}. \\[2ex]
  I_R & \geq &
  \min_{\vu,\v{\check\u}}[H_{R\,left}\!+\!H_{R\,right}\!-\!H_{R\,joint}]
  \\
  & \geq & H_{R\,left}^{lb}\!+\!H_{R\,right}^{lb}\!-\!H_{R\,joint}^{ub}
  = H_{R\,left}^{lb}\!+\!H_{R\,right}^{lb} =: I_R^{lb}.
\eqan
Note that for $H_R$ we can tolerate such a crude approximation,
since $H_R$ (and $H_R^{ub/lb}$) are small $O(\sigma^2)$
corrections. In summary we have
\bqan
  \overbrace{I(\upa\vu)}^{\up I-O(\sigma^2)} & \leq &
  \up I \;\leq\; \overbrace{I_0 + \up{I_1}}^{\up I+O(\sigma^2)}
  +\overbrace{I_R^{ub}}^{O(\sigma^2)}
\qmbox{and} \\[2ex]
  \underbrace{I_0 + \low{I_1}}_{\low I-O(\sigma^2)} +
  \underbrace{I_R^{lb}}_{O(\sigma^2)}
  & \leq & \low I \;\leq\;
  \underbrace{I(\lowa\vu)}_{\low I+O(\sigma^2)}.
\eqan

%-------------------------------------------------------------%
\subsection{Comparing edges}\label{secCE}
%-------------------------------------------------------------%

For two edges $a$ and $b$ with no common vertex, the reliable
interval containing $[\low I,\up I]$ of Section~\ref{secREMI} can
be used separately for $a$ and $b$. For edges with a common vertex
the results of Section~\ref{secREMI} may still be used, but they
may no longer be reliable or good from a global perspective.
Consider the subgraph
$\imath\stackrel{a}{\mbox{---}}\jmath\stackrel{b}{\mbox{---}}\kappa$,
joint probabilities $\pi_{\imath\jmath\kappa}$ of vertices
$\imath$, $\jmath$, $\kappa$, a Dirichlet posterior
$\prod_{ij\kdot}\pi_{ij\kdot}^{n_{ij\kdot}+st_{ij\kdot}-1}$,
$\u_{ij\kdot}=E_\vt[\pi_{ij\kdot}]={n_{ij\kdot}+st_{ij\kdot}\over
n+s}$, etc. The expected mutual information between node $\imath$
and $\jmath$ is $I^a:=I(\vu^a)$ and $I^b:=I(\vu^b)$ between $\jmath$
and $\kappa$, where $\u^a_{ij}=\u_{ij\p}$ and
$\u^b_{j\kdot}=\u_{\p j\kdot}$. The weight of edge $a$ is $w^a=
[\min\, I^a,\max\,I^a]$, where $\min$ and $\max$ are w.r.t.\
$t^a_{ij}:=t_{ij\p}$. Similarly, the weight of edge $b$ is $w^b =
[\min\,I^b ,\max\,I^b]$, where $\min$ and $\max$ is w.r.t.\
$t^b_{j\kdot}:=t_{\p j\kdot}$. The results of
Section~\ref{secREMI} can be used to determine the intervals.
Unfortunately this procedure neglects the constraint $t^a_{\p
j}=t^b_{j\p}$. The correct treatment is to define $w^a$ larger
than $w^b$ as follows:
\beqn
  [w^a > w^b] \quad\Leftrightarrow\quad
  [I^a > I^b \;\mbox{for all}\; t_{\imath\jmath\kappa}\in\Delta]
  \quad\Leftrightarrow\quad
  \min_\vt[I^a-I^b]>0.
\eeqn
The crude approximation $\min[I^a-I^b]\geq\min I^a-\max I^b$ gives
back the above naive interval comparison procedure. This shows
that the naive procedure is reliable, but the approximation may be
crude. For good estimates we proceed similar as in
Section~\ref{secREMI} to get $O(\sigma^2)$ approximations and
bounds on $I^a-I^b$.
\beqn
  \overbrace{I^a_0\!-\!I^b_0\!+\!
  I^a_1(\lowa\vu)\!-\!I^b_1(\lowa\vu)}^{\min[I^a-I^b]-O(\sigma^2)}
  +\overbrace{I^{a.lb}_R-I^{b.ub}_R}^{O(\sigma^2)}
  \leq \min_{\vt\in\Delta}[I^a\!-\!I^b] \leq
  \overbrace{I^a(\lowa\vu)\!-\!I^b(\lowa\vu)}^{\min[I^a-I^b]+O(\sigma^2)}
\eeqn
\beqn
  \lowa{(ij\kdot)}:=\argmin_{ij\kdot}
  [h'(\u_{i\pp}^*)-h'(\u_{ij\p}^*)-h'(\u_{\pp\kdot}^*)+h'(\u_{\p j\kdot}^*)]
\eeqn
\beqn
  =\arg\limits_{(ij\kdot)}\{\min_j[\min_i(h'(\u_{i\pp}^*)-h'(\u_{ij\p}^*))+
                     \min_\kdot(h'(\u_{\p j\kdot}^*)-h'(\u_{\pp\kdot}^*))]\}
\eeqn
and $\lowa t_{ij\kdot}:=\delta_{(ij\kdot),\lowa{(ij\kdot)}}$, and,
for instance, choosing $t_{ij\kdot}^*={1\over d_\imath d_\jmath
d_\kappa}$ or $t_{ij\kdot}^*={n_{ij\kdot}\over n}=\u_{ij\kdot}^*$.
The second representation for $\lowa{(ij\kdot)}$ shows that
$\lowa{(ij\kdot)}$, and hence the bounds, can be computed in time
$O(d^2)$ rather than $O(d^3)$. Note that $\min_i$ and $\min_\kdot$
determine $i$ and $\kdot$ as a function of $j$, then $\min_j$
determines $\lowa j$, which can be used to get $\lowa i=i(\lowa
j)$ and $\lowa\kdot=\kdot(\lowa j)$. This lower bound on
$\min[I^a-I^b]$ is used in the next section to robustly compare
weights.

%%%%%%%%%%%%%%%%%%%%%%%%%%%%%%%%%%%%%%%%%%%%%%%%%%%%%%%%%%%%%%%
\section{An example}\label{secAE}
%%%%%%%%%%%%%%%%%%%%%%%%%%%%%%%%%%%%%%%%%%%%%%%%%%%%%%%%%%%%%%%

This section illustrates the application of the developed
methodology to an artificial problem.

\begin{figure}[h]\begin{center}
  \includegraphics[width=8cm]{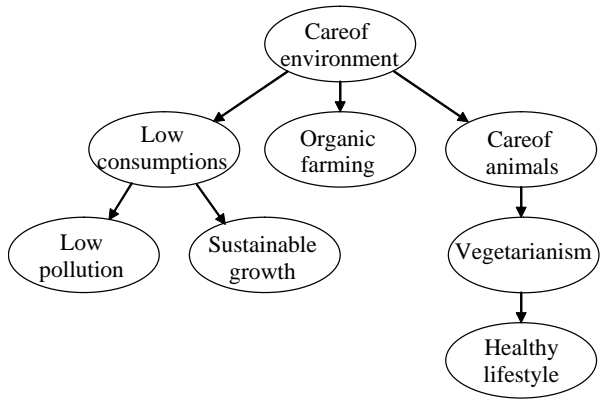}\\
  \caption{A graph that models the dependencies between the
  random variables of an artificial domain.}\label{fig30}
\end{center}\end{figure}

The graph in Figure~\ref{fig30} models the domain by relationships
of direct dependency, represented by directed arcs. Each node
represents a binary (yes-no) variable that is associated with the
probability distribution of the variable itself conditional on the
state of the parent node. The distributions are given in Table~\ref{tab10}.

\begin{table}[h]
  \centering
  \begin{tabular}{l c c}
    \hline
      Variable & P(variable=yes$\vert$parent=yes) & P(variable=yes$\vert$parent=no) \\
      \hline
      Care of environment & 0.366 & 0.366 \\
      Low consumptions & 0.959 & 0.460 \\
      Organic farming & 0.950 & 0.450 \\
      Care of animals & 0.801 & 0.332 \\
      Low pollution & 1.000 & 0.208 \\
      Sustainable growth & 0.951 & 0.200 \\
      Vegetarianism & 0.993 & 0.460 \\
      Healthy lifestyle & 0.920 & 0.300 \\
      \hline
  \end{tabular}
  \caption{Conditional probability distributions for the variables of the example in Figure~\ref{fig30}. %
  (The distribution of `Care of environment' is represented in this table %
  though it is actually unconditional.)}\label{tab10}
\end{table}

A model made by the graph and the probability tables, as the one
above, is called a \emph{Bayesian network} \cite{Pearl88}. We used
the Bayesian network to sample units from the joint distribution
of the variables in the graph. Each unit is a vector that
represents a joint instance of all the variables. By the generated
data set we can test our algorithm for the discovery of strong
edges, and compare it with Chow and Liu's algorithm.

The `strong edges algorithm' is summarized for clarity in
Table~\ref{strong edges}. The main procedure is called
`DetectStrongEdges' and it implements the exact procedure from
Section~\ref{secEDOSE}. The comparison of edges needed by
`DetectStrongEdges' is implemented by the subprocedure
`TestDominance'. The test~\ref{testnocommonnode} there exploits
the bounds defined in Section~\ref{secREMI} (we have added
superscripts $a$ and $b$ to the terms of the bounds to make it
clear to which edge they refer). For edges with a common node, the
test~\ref{testwithcommonnode} exploits the bounds given in
Section~\ref{secCE}. For the dominance tests we have used the
value 1 for the IDM hyper-parameter $s$ (see
Section~\ref{secIDM}). We have also chosen
$t_{ij}^*=\frac{1}{d_\imath d_\jmath}$,
$t_{ij\kdot}^*=\frac{1}{d_\imath d_\jmath d_\kappa}$, etc.

\begin{table}

\begin{enumerate}
    \item Procedure \textbf{DetectStrongEdges}(a set-based weighted graph $G$)
    \begin{enumerate}
        \item forest:=$\emptyset$;
        \item for each edge $e\in E$
        \begin{enumerate}
            \item consider $G'$ obtained from $G$ dropping $e$ and the edges it dominates;
            \item if the endpoints of $e$ are not connected in $G'$, add $e$ to forest;
        \end{enumerate}
        \item return forest.
    \end{enumerate}

    \item Procedure \textbf{TestDominance}(edge $a$, edge $b$)
    \begin{enumerate}
        \item if $a$ and $b$ do not share nodes then (i.e., the edges are $\imath\stackrel{a}{\mbox{---}}\jmath$ and $\tilde{\imath}\stackrel{b}{\mbox{---}}\tilde{\jmath}$)
        \begin{enumerate}
            \item $I^a_0:=\sum_{i} h(\u_{i+}^*)+\sum_j h(\u_{+j}^*)-\sum_{ij} h(\u_{ij}^*)$;
            \item $I^b_0:=\sum_{\tilde{i}} h(\u_{\tilde{i}+}^*)+\sum_{\tilde{j}}h(\u_{+\tilde{j}}^*)-\sum_{\tilde{i}\tilde{j}} h(\u_{\tilde{i}\tilde{j}}^*)$;
            \item $\low{I_1^a}:=\;\;\;\sigma\,\min_{ij}[h'(\u_{i+}^*)+h'(\u_{+j}^*)-h'(\u_{ij}^*)]$ \\
                   \hspace*{5ex} $-\sigma\sum_{ij}t_{ij}^*[h'(\u_{i+}^*)+h'(\u_{+j}^*)-h'(\u_{ij}^*)]$;
            \item $\up{I_1^b}:=\;\;\;\sigma\,\max_{\tilde{i}\tilde{j}}[h'(\u_{\tilde{i}+}^*)+h'(\u_{+\tilde{j}}^*)-h'(\u_{\tilde{i}\tilde{j}}^*)]$ \\
                   \hspace*{5ex} $-\sigma\sum_{\tilde{i}\tilde{j}}t_{\tilde{i}\tilde{j}}^*[h'(\u_{\tilde{i}+}^*)+h'(\u_{+\tilde{j}}^*)-h'(\u_{\tilde{i}\tilde{j}}^*)]$;
            \item $I^{a.lb}_R:=\frac{1}{2}\sigma^2\sum_{i}h''(\frac{n_{i+}}{n+s}) + \frac{1}{2}\sigma^2\sum_{j}h''(\frac{n_{+j}}{n+s})$;
            \item $I^{b.ub}_R:=-\frac{1}{2}\sigma^2\sum_{\tilde{i}\tilde{j}}h''(\frac{n_{\tilde{i}\tilde{j}}}{n+s})$;
            \item\label{testnocommonnode}if $I_0^a-I_0^b + \low{I_1^a}-\up{I_1^b}+I_R^{a.lb}-I_R^{b.ub}>0 $, return `true';
        \end{enumerate}
        \item else (i.e., the edges are $\imath\stackrel{a}{\mbox{---}}\jmath\stackrel{b}{\mbox{---}}\kappa$)
        \begin{enumerate}
            \item $I^a_0:=\sum_{i} h(\u_{i++}^*)+\sum_j h(\u_{+j+}^*)-\sum_{ij} h(\u_{ij+}^*)$;
            \item $I^b_0:=\sum_{j} h(\u_{+j+}^*)+\sum_\kdot h(\u_{++\kdot}^*)-\sum_{j\kdot} h(\u_{+j\kdot}^*)$;
            \item $\low{I_1^a-I_1^b}:=\sigma\,\min_j[\min_i(h'(\u_{i\pp}^*)-h'(\u_{ij\p}^*))
                                             +\min_\kdot(h'(\u_{\p j\kdot}^*)-h'(\u_{\pp\kdot}^*))]$ \\
              $-\sigma\sum_j[d_\kappa\sum_i t_{ij\kdot}^*(h'(\u_{i\pp}^*)-h'(\u_{ij\p}^*))
                                          +d_\imath\sum_\kdot t_{ij\kdot}^*(h'(\u_{\p j\kdot}^*)-h'(\u_{\pp\kdot}^*))]$;
            \item $I^{a.lb}_R:=\frac{1}{2}\sigma^2\sum_{i}h''(\frac{n_{i++}}{n+s}) + \frac{1}{2}\sigma^2\sum_{j}h''(\frac{n_{+j+}}{n+s})$;
            \item $I^{b.ub}_R:=-\frac{1}{2}\sigma^2\sum_{j\kdot}h''(\frac{n_{+j\kdot}}{n+s})$;
            \item\label{testwithcommonnode} if $I^a_0-I^b_0+\low{I_1^a-I_1^b}+I^{a.lb}_R-I^{b.ub}_R>0$, return `true';
        \end{enumerate}
        \item return `false'.
    \end{enumerate}
    \item Procedure \textbf{h}($u$) return $u\psi(n+s+1)-u\psi(nu+su+1)$;
    \item Procedure \textbf{h}$\mathbf'$($u$) return $\psi(n+s+1)-\psi(nu+su+1)-u(n+s)\psi'(nu+su+1)$;
    \item Procedure \textbf{h}$\mathbf{''}$($u$) return $-2(n+s)\psi'(nu+su+1)-u(n+s)^2\psi''(nu+su+1)$;
\end{enumerate}

\caption{A summary view of the strong edges algorithm. Remember
that $\sigma=\frac{s}{n+s}$, $n$ is the sample size,
$u_{\cdots}=\frac{n_{\cdots}+t_{\cdots}}{n+s}$ denotes the
expectation of a certain chance, $u_{\cdots}^*$ the expectation
taken for a specific value $t_{\cdots}^*$ of hyper-parameter
$t_{\cdots}$; finally, $\psi$ denotes the $\psi$-function,
described in Appendix~\ref{secPsi}.}\label{strong edges}
\end{table}

\begin{figure}[H]
\ifjournal\tabcapfont\fi
\centerline{%
\begin{tabular}{c@{\hspace{0.5cm}}c}
\includegraphics[width=5.5cm]{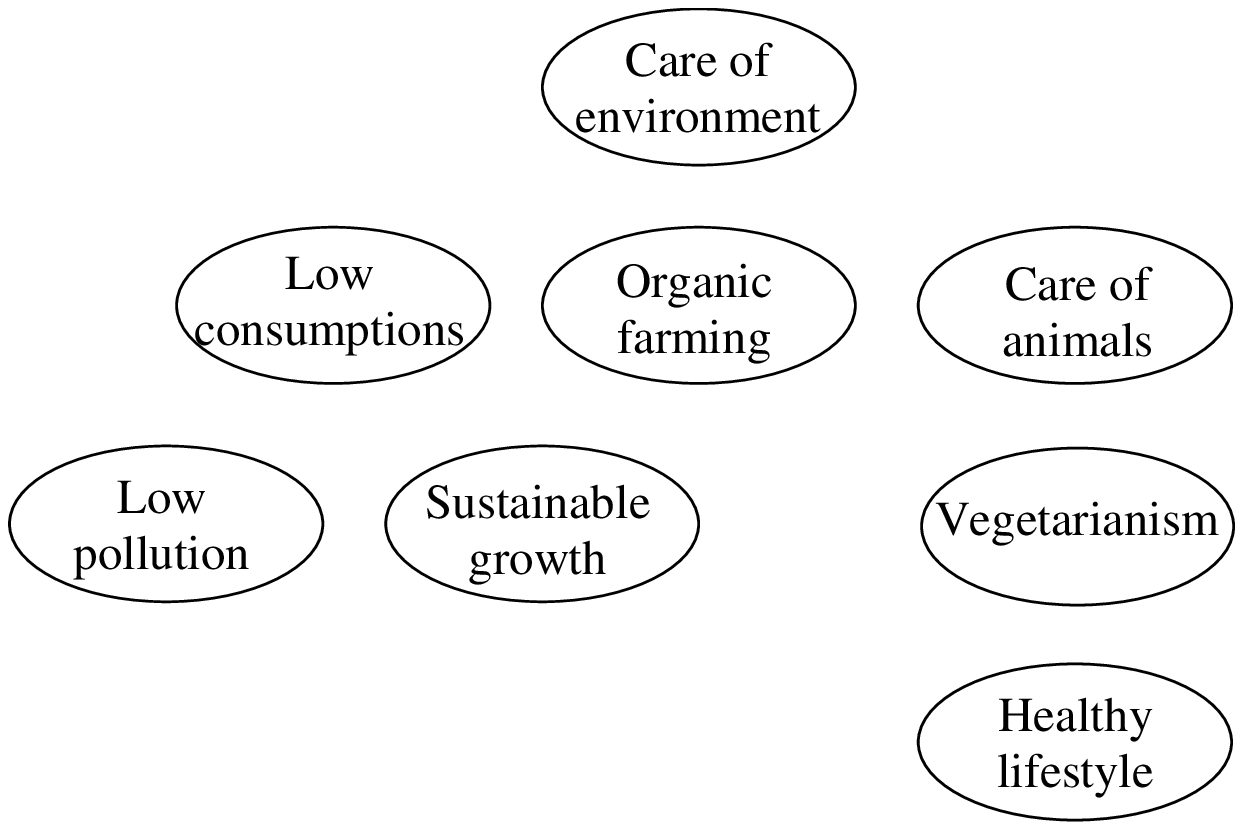} &
\includegraphics[width=5.5cm]{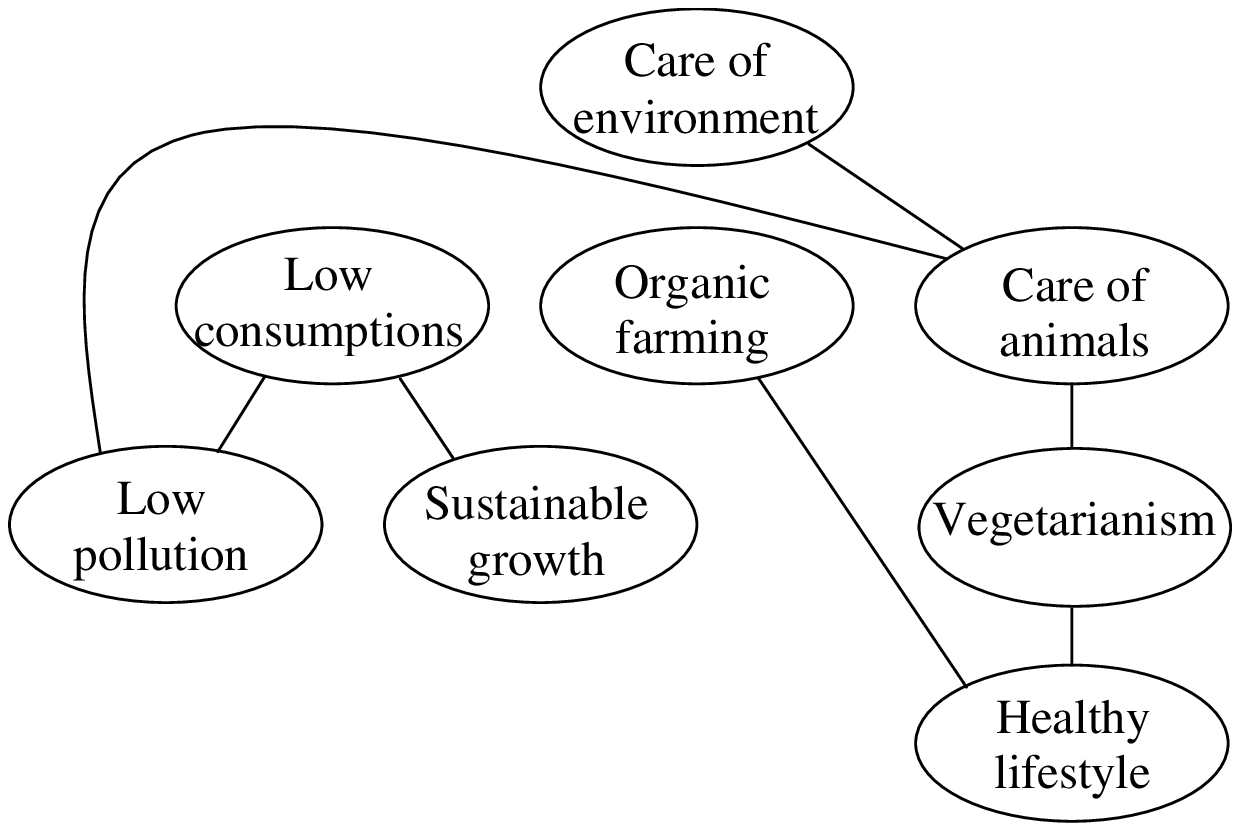} \\
a.~~ Strong edges algorithm & b.~~ Chow and Liu's algorithm
\end{tabular}}
\caption{The outputs of the two algorithms after reading 20 instances.}\label{fig40}
\end{figure}

\begin{figure}[H]
\ifjournal\tabcapfont\fi
\centerline{%
\begin{tabular}{c@{\hspace{0.5cm}}c}
\includegraphics[width=5.5cm]{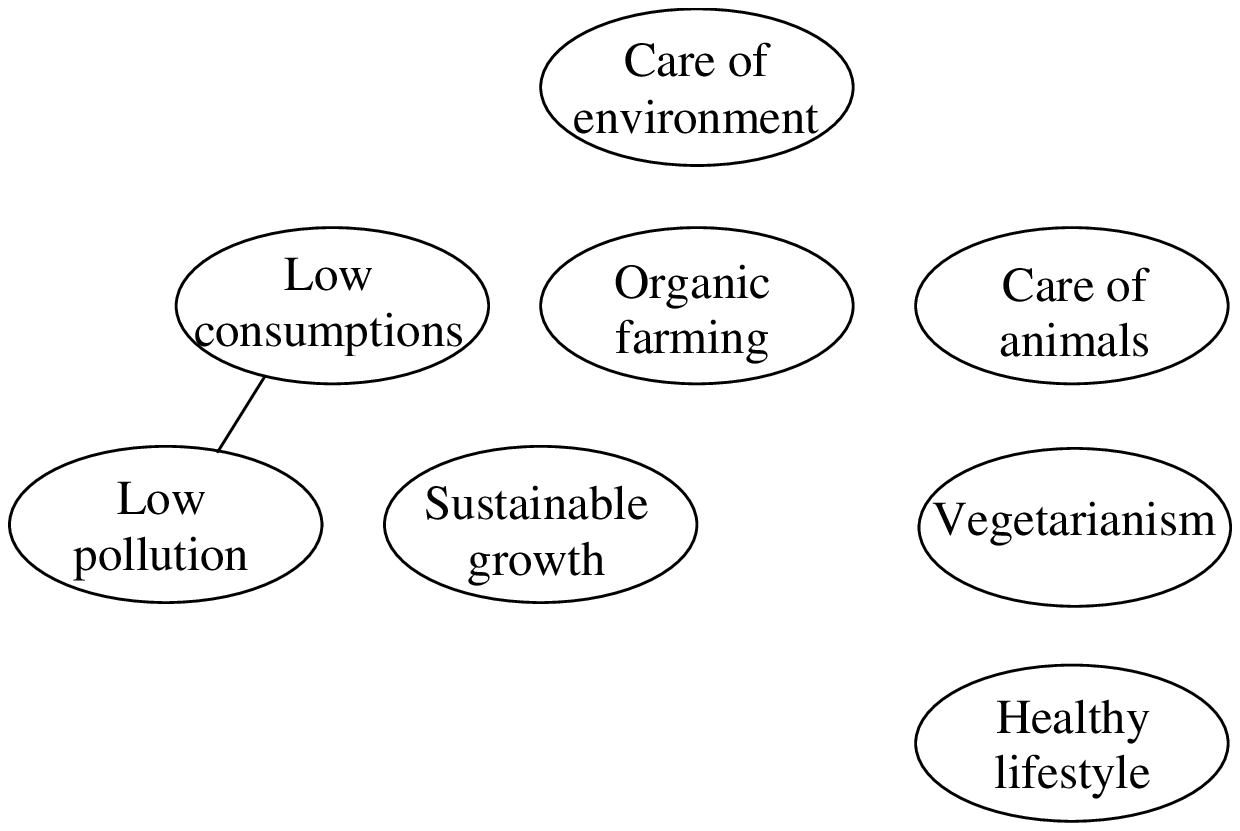} &
\includegraphics[width=5.5cm]{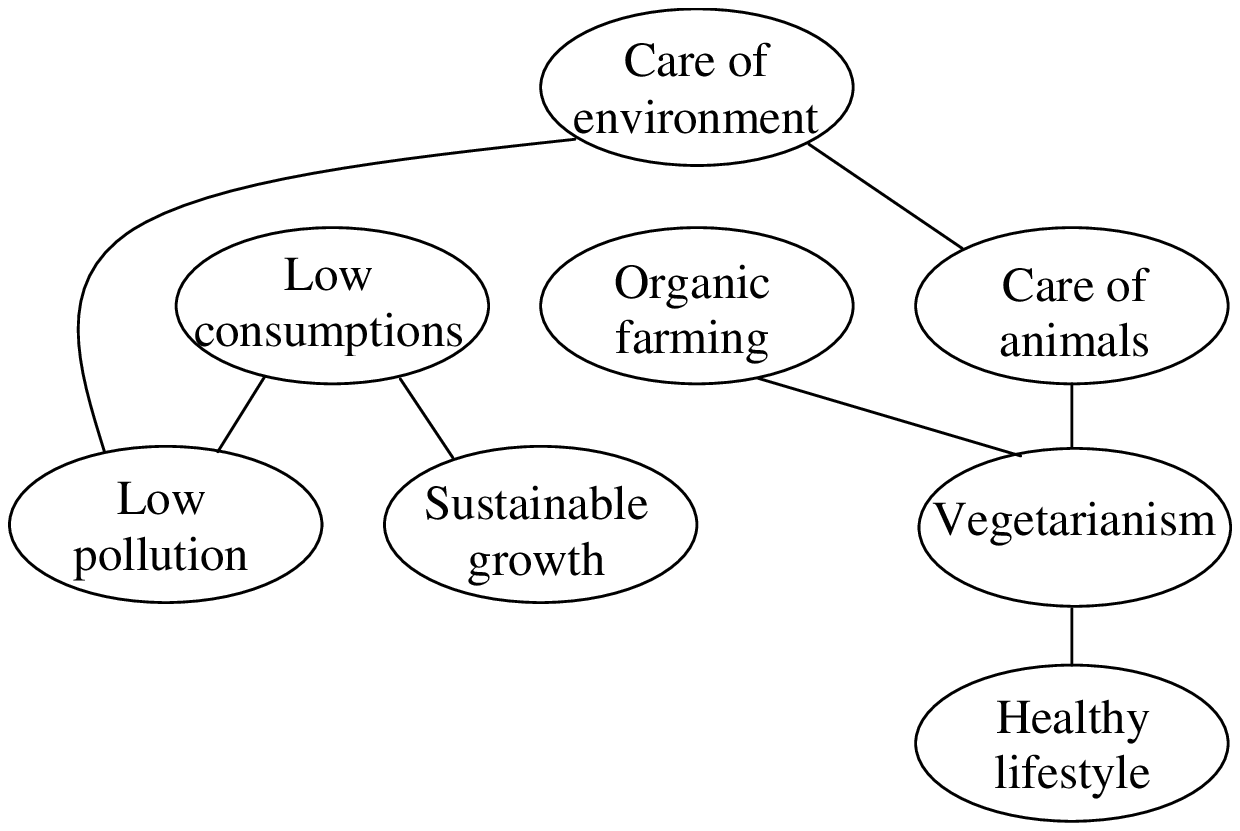} \\
a.~~ Strong edges algorithm & b.~~ Chow and Liu's algorithm
\end{tabular}}
\caption{The outputs of the two algorithms after reading 30 instances.}\label{fig50}
\end{figure}

\begin{figure}[H]
\ifjournal\tabcapfont\fi
\centerline{%
\begin{tabular}{c@{\hspace{0.5cm}}c}
\includegraphics[width=5.5cm]{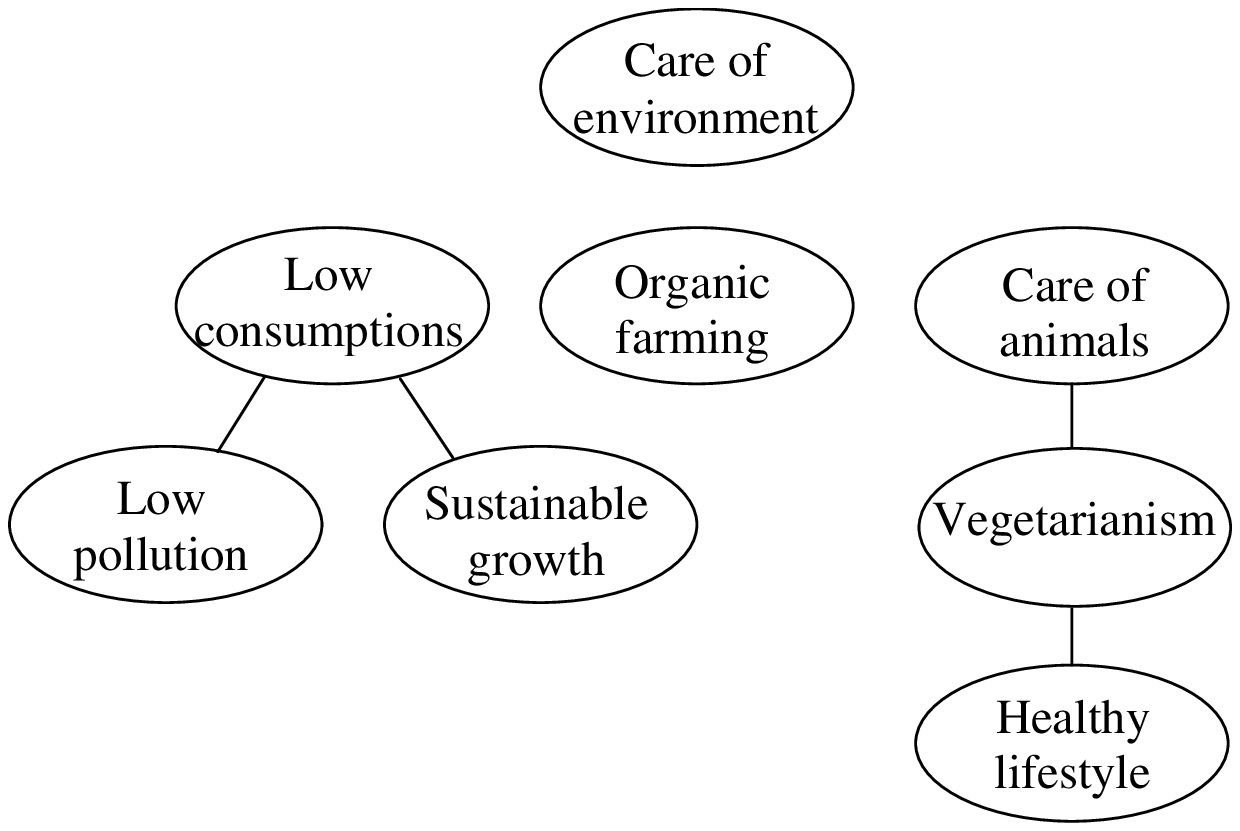} &
\includegraphics[width=5.5cm]{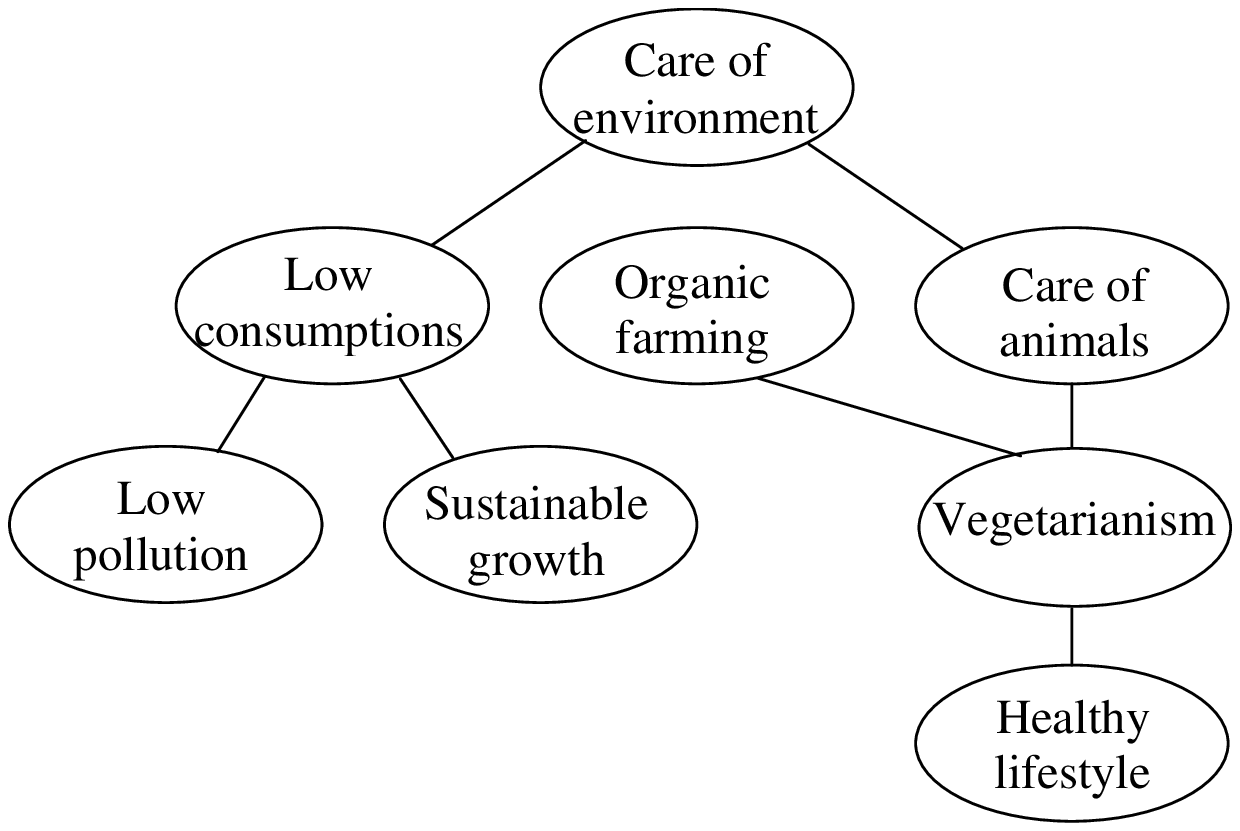} \\
a.~~ Strong edges algorithm & b.~~ Chow and Liu's algorithm
\end{tabular}}
\caption{The outputs of the two algorithms after reading 40 instances.}\label{fig60}
\end{figure}

\begin{figure}[H]
\ifjournal\tabcapfont\fi
\centerline{%
\begin{tabular}{c@{\hspace{0.5cm}}c}
\includegraphics[width=5.5cm]{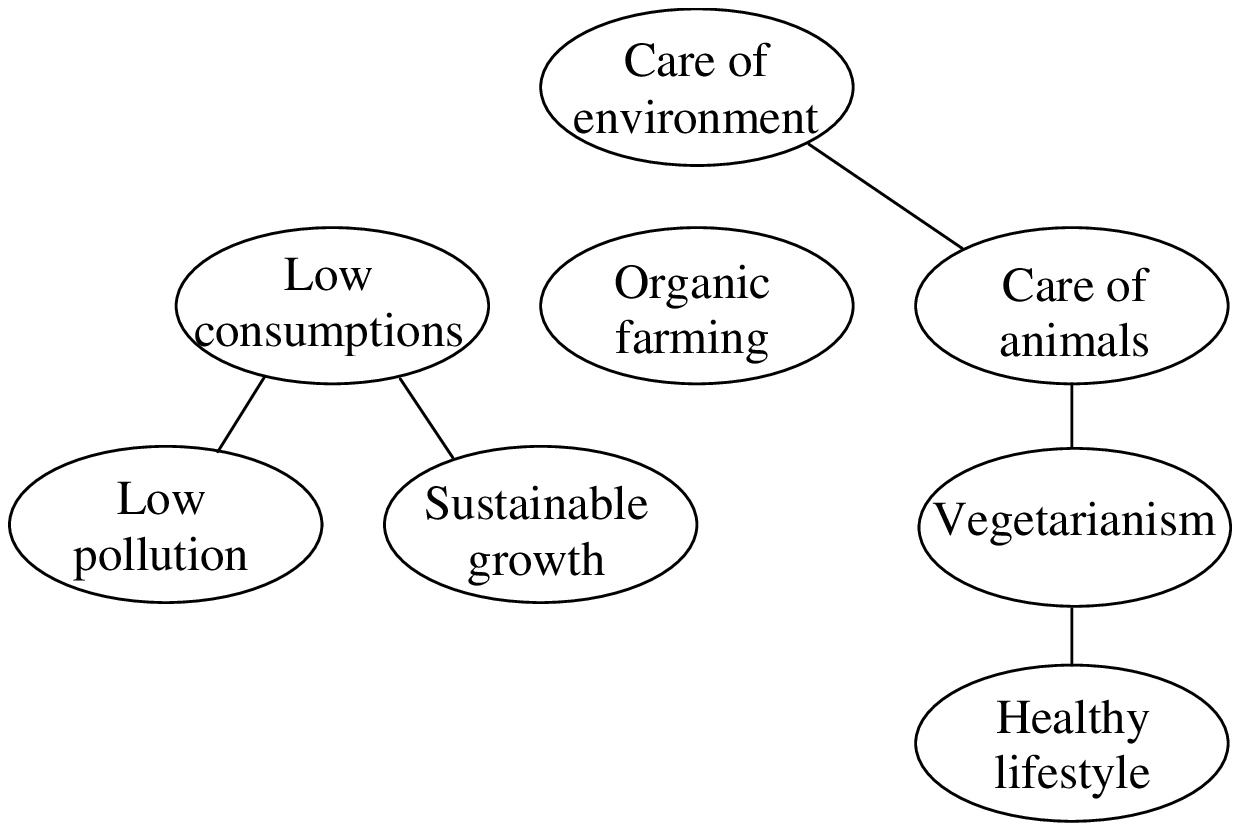} &
\includegraphics[width=5.5cm]{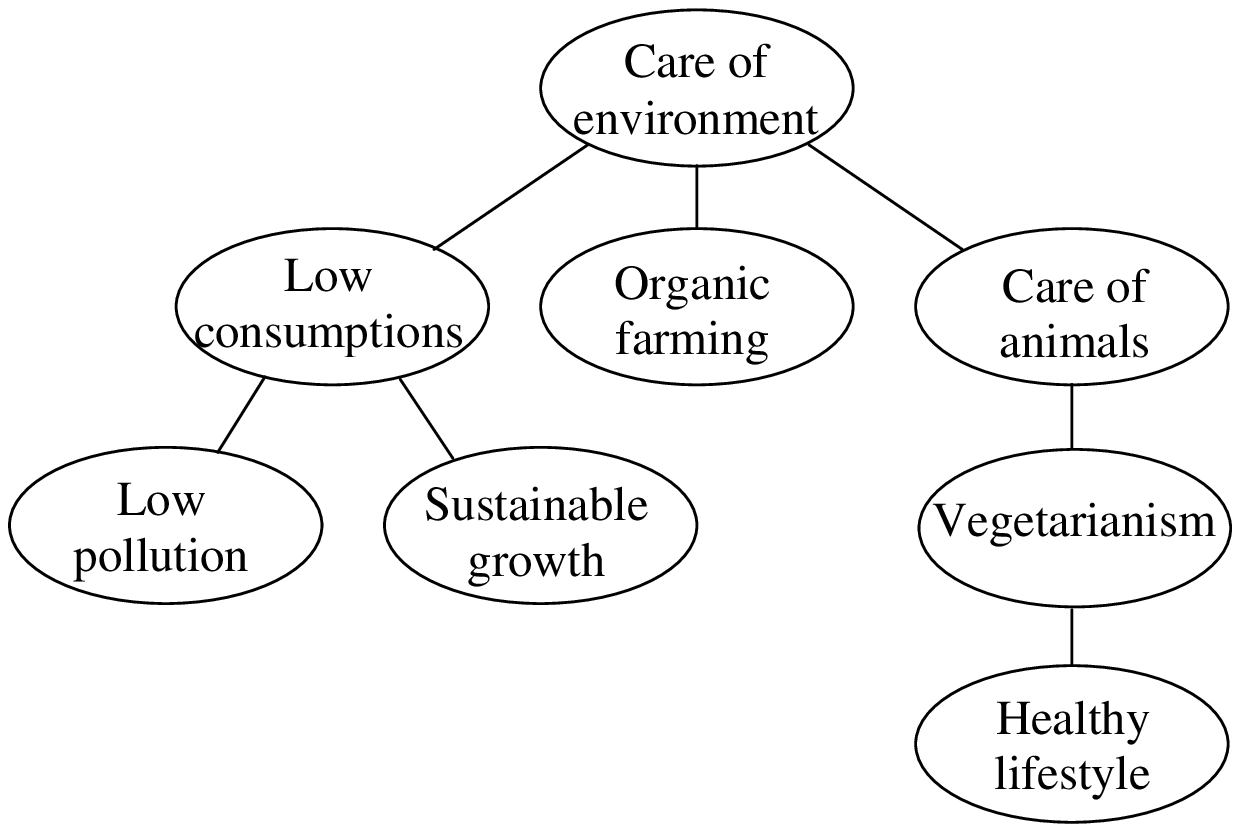} \\
a.~~ Strong edges algorithm & b.~~ Chow and Liu's algorithm
\end{tabular}}
\caption{The outputs of the two algorithms after reading 50 instances.}\label{fig70}
\end{figure}

Figures~\ref{fig40} to~\ref{fig70} show the progression of the
models discovered by the two algorithms as more instances are
read. The strong edges algorithm appears to behave more reliably
than Chow and Liu's algorithm. It suspends the judgment on
ambiguous cases and outputs forests. These are always composed of
edges of the actual graph. Chow and Liu's algorithm always
produces complete trees, but these misrepresent the actual tree
until 50 instances have been read. At this point Chow and Liu's
algorithm detects the right tree. The cautious approach
implemented by the strong edges algorithm needs other 20 instances
to produce the same complete tree.

%%%%%%%%%%%%%%%%%%%%%%%%%%%%%%%%%%%%%%%%%%%%%%%%%%%%%%%%%%%%%%%
\section{Extensions}\label{secE}
%%%%%%%%%%%%%%%%%%%%%%%%%%%%%%%%%%%%%%%%%%%%%%%%%%%%%%%%%%%%%%%

The methodology developed so far leads naturally to other possible
extensions of Chow and Liu's approach. We briefly report on two
different types of extensions in the following.

Section~\ref{secRT} discusses the question of tree-dependency
structures vs. forest-dependency structures under several
respects. The discussion focuses both on algorithms that are
alternative to the strong edges one, and that aim at yielding
trees, and on the other hand on algorithms that emphasize the
inference of forest-dependency structures from data.

In Section~\ref{secCLFEAMI} we extend the computation of lower and
upper expectations of mutual information to the computation of
robust credible limits. These are intervals for mutual information
obtained from the IDM that contain the actual value with given
probability. This result is useful in order to produce dependency
structures that provide the user with a given guarantee level. In
principle the extension to credible limits can be applied both to
the computation of strong edges and to that of robust trees, as
defined in the next section, although the results of
Sections~\ref{secRT} and~\ref{secCLFEAMI} are actually
independent, in the sense that one does not need to use them
together.

%-------------------------------------------------------------%
\subsection{Forests vs. Trees}\label{secRT}
%-------------------------------------------------------------%

It may be useful to critically re-consider Chow and Liu's
algorithm in the following respect. Chow and Liu's algorithm
yields always a tree by construction, and hence this happens also
when the actual (but usually unknown) dependency structure is a
forest. This is a questionable characteristic of the algorithm, as
in the mentioned case yielding a tree seems to be hard to justify.
There are indeed approaches in the literature of precise
probability that suppress the edges of a maximum spanning tree for
which the mutual information is not large enough, yielding a
forest. This is typically implemented using a numerical threshold
$\varepsilon$, sometimes computed via statistical tests. Such
approaches can be used immediately also within the
imprecise-probability framework introduced in this paper; it is
sufficient to suppress the edges for which the upper value of
mutual information [i.e., $\max_{w\in W}w(e)$] does not exceed
$\varepsilon$. In contrast with the precise-probability approach,
the latter should have the advantage to better deal with the
problem to suppress edges by mistake, as a consequence of the
variability of the inferred values of mutual information. This
should be especially true once forests are inferred using the
credible limits for mutual information introduced in the next
section.

A more subtle question is how the forests inferred using the above
threshold procedure relate to the forests that are naturally
produced by the strong edges algorithm in its original form.
Remind that the strong edges algorithm produces a forest rather
than a tree when there is more than one optimal tree consistent
with the available data; indeed the algorithm aims at yielding the
graphical structure made of the intersection of all such trees.
The situation may be clarified by focusing on a special case:
consider a problem in which the true dependency structure is a
tree in which there are edges with the same value of mutual
information, say $\mu$. In this case the strong edges algorithm
will never produce a tree, only a forest, also in the limit of
infinitely many data. The reason is that there will always be
multiple optimal trees consistent with the data, just because
multiple optimal trees are a characteristic of the problem. In
particular, there would arise a forest because some edges with
weight $\mu$ would never belong to the set of strong edges. Now suppose
that $\mu>\varepsilon$. In this case, the previous threshold
procedure would not suppress the edges with mutual information
equal to $\mu$. In other words, the two procedures suppress edges
under different conditions: the strong edges algorithm may
suppress edges because they have equal true values of mutual
information, despite those values may be high; the threshold
procedure only suppresses edges with low value of mutual
information. For this reason it could make sense to apply the
threshold procedure also as a post-processing step of the strong
edges algorithm.

The discussion so far has highlighted an interesting point. By
focusing on the intersection of all the trees consistent with the
data, the strong edges algorithm appears to be well suited as a
tool to recover the actual dependency structure underlying the
data. This is because the algorithm does not aim at recovering
just any of the equivalent structures, rather, it focuses on the
common pattern to all of them, which is obviously part of the
actual structure. In this sense, the strong edges algorithm might
be well suited for applications concerned with the recovery of
causal patterns.

On the other hand, one can think of applications for which the
algorithm is probably not so well suited. For instance, in
(precise-probability) problems of pattern classification based on
Bayesian networks \cite{FrGeGo97}, it is important to recover any
tree (or forest) structure for which the sum of the edge weights
is maximized. In this case, suppressing edges with large weights
only because they are not strong might lead to low classification accuracy. In these cases,
the extension of those precise approaches to the IDM-based
inferential approach should probably follow other lines than those
described here. One possibility could be to exploit existing
results in the literature of robust optimization; the work of
Yaman et al. \cite{YamKarPin01} seems to be particularly worthy of
consideration. Yaman et al. consider a problem of maximum spanning
tree for a graph with weights specified by intervals (the weights
are given no particular interpretation), which is a special case
of a set-based weighted graph. They define the \emph{relative
robust spanning tree} as follows (using our notations): let $T$ be
a generic tree spanning $G$, and denote by $T^*_w$ a maximum
spanning tree of $G_w\in\mathcal{G}$. Let $S^*_w$ resp. $S_w$ be
the sum of the edge weights of $T^*_w$ resp. $T$, with respect to
the weight function $w$. A relative robust spanning tree $T^*$ is
one that solves the optimization problem $\min_{T}\max_{w\in
W}(S_{w}^{\ast }-S_{w})$, i.e., one that minimizes the largest
deviation $S_{w}^{\ast }-S_{w}$ among all the possible graphs
$G_w\in\mathcal{G}$. In this sense the approach adopted by Yaman
et al. is in the long tradition of the popular \emph{maximin} (or
\emph{minimax}) decision criterion. From the computational point
of view, although the problem is \emph{NP-complete}
\cite{aron2004}, recent results show that relatively large
instances of the problem can be solved efficiently
\cite{montemanni2005a}. The trees defined by Yaman et al. could
probably be combined with the IDM-based inferential approach
presented here, suitably modified for classification problems, in
order to yield relative robust classification trees. Here, too, it
could make sense to post-process the relative robust trees in
order to suppress edges with small upper (or even lower) values of
mutual information, yielding a forest.

%-------------------------------------------------------------%
\subsection{Robust credible limits for mutual information}\label{secCLFEAMI}
%-------------------------------------------------------------%

In this section we develop a full inferential approach for mutual
information under the IDM.

An $\alpha$-credible interval for the mutual information $\cal I$
is an interval $[\undertilde{\cal I},\widetilde{\cal
I}]$ which contains $\cal I$ with probability at least $\alpha$,
i.e., \ $\int_{\undertilde{\cal I}}^{\widetilde{\cal I}}
p({\cal I})d{\cal I}\geq\alpha$. We define $\alpha$-credible
intervals w.r.t.\ distribution $p_\vt({\cal I})$ as
\beqn
%  {\mathop{\widetilde{\cal I}}\limits_\sim}\!\,_\vt \;=\;
  [ \undertilde{\cal I}_\vt , \widetilde{\cal I}_\vt] \;=\;
  [ E_\vt[{\cal I}]-\undertilde{\Delta\cal I}_\vt \,,\,
  E_\vt[{\cal I}]+\widetilde{\Delta\cal I}_\vt ]
  \qmbox{such that}
  \int_{\undertilde{\cal I}_\vt}^{\widetilde{\cal I}_\vt}
  p_\vt({\cal I})d{\cal I}\geq\alpha,
\eeqn
where
$\widetilde{\Delta\cal I}_\vt:=\widetilde{\cal I}_\vt-E_\vt[{\cal I}]$
($\undertilde{\Delta\cal I}_\vt :=
E_\vt[{\cal I}]-\undertilde{\cal I}_\vt$) is the distance from the
right boundary $\widetilde{\cal I}_\vt$ (left boundary
$\undertilde{\cal I}_\vt$) of the $\alpha$-credible
interval
%$\undertilde{\widetilde{\cal I}}_\vt$
$[\undertilde{\cal I}_\vt,\widetilde{\cal I}_\vt]$
to the mean
$E_\vt[{\cal I}]$ of $\cal I$ under distribution $p_\vt$. We can use
\beqn
  [\underbartilde{\cal I} , \up{\widetilde{\cal I}}] \;:=\;
  [ \min_\vt\undertilde{\cal I}_\vt , \max_\vt\widetilde{\cal I}_\vt]
  \;=\; \bigcup_\vt\,
  [ \undertilde{\cal I}_\vt , \widetilde{\cal I}_\vt]
\eeqn
as a robust credible interval, since
$\int_{\underbartilde{\cal I}}^{\up{\widetilde{\cal I}}}
  p_\vt({\cal I})d{\cal I} \geq \int_{\undertilde{\cal I}_\vt}^{\widetilde{\cal I}_\vt}
  p_\vt({\cal I})d{\cal I} \geq \alpha$ for all
$\vt$. An upper bound for $\up{\widetilde{\cal I}}$ (and
similarly lower bound for $\underbartilde{\cal I}$) is
\beqn
  \up{\widetilde{\cal I}} \;=\;
  \max_\vt(E_\vt[{\cal I}]+\widetilde{\Delta\cal I}_\vt)
  \;\leq\;
  \max_\vt E_\vt[{\cal I}]+\max_\vt \widetilde{\Delta\cal I}_\vt
  \;=\;   \up{E[{\cal I}]} + \up{\widetilde{\Delta\cal I}}.
\eeqn
Good upper bounds on $\up I= \up{E[{\cal I}]}$ have been
derived in Section~\ref{secREMI}.

For not too small $n$, $p_\vt({\cal I})$ is close to Gaussian due
to the central limit theorem. So we may approximate
$\widetilde{\Delta\cal I}_\vt\approx r\sigma_\vt$ with $r$ given
by $\alpha=\mbox{erf}(r/\sqrt{2})$, where erf is the error
function (e.g., \ $r=2$ for $\alpha\approx 95\%)$ and $\sigma_\vt$
is the variance of $p_\vt$, keeping in mind that this could be a
non-conservative approximation. In order to determine
$\up{\widetilde{\Delta\cal I}}$ we only need to estimate
$\max_\vt\sqrt{\Var_\vt[{\cal I}]}=O({1\over n})$.
The variation of $\sqrt{\Var_\vt[{\cal I}]}$ with $\vt$ is
of order $n^{-3/2}$. If we regard this as negligibly small, we may
simply fix some $\vt^*\in\Delta$.
So the robust credible interval for $\cal I$ can be
estimated as
\beqn
  \up{\widetilde{\cal I}}
  \;\leq\; \up I+\up{\widetilde{\Delta\cal I}}
  \;\leq\; I_0+\up{I_1}+I_R^{ub}+\up{\widetilde{\Delta\cal I}}
  \;\approx\; I_0+\up{I_1}+I_R^{ub}+r\sqrt{\Var_{\vt^*}[{\cal I}]}.
\eeqn
Expressions for the variance of $\cal I$ have been
derived in \cite{Hut01,HutZaf05}:
\beqn
  \Var_\vt[{\cal I}] =
  {1\over n\!+\!s}
  \sum_{ij}\u_{ij}\bigg(\log{\u_{ij}\over
    \u_{i\p}\u_{\p j}}\bigg)^2 \!-\!
  {1\over n+s}\bigg(\sum_{ij}\u_{ij}\log{\u_{ij}\over
    \u_{i\p}\u_{\p j}}\bigg)^2 \!\!+ O(n^{-2}).
\eeqn
Higher order corrections to the variance and higher moments have
also been derived, but are irrelevant in light of our other
approximations. In Sections~\ref{secCE} and~\ref{secAE} we also
needed a lower bound on $I^a-I^b$. Taking credible intervals into
account we need a robust upper $\alpha$-credible limit for ${\cal
I}^{ba}:={\cal I}^b-{\cal I}^a$. Similarly as for the variance one
can derive the following expression:
\bqan
  \up{\widetilde{{\cal I}^{\smash{ba}}}} & \leq &
  I^b_0-I^a_0+\up{I^b_1}-\low{I^a_1}+I^{b.ub}_R-I^{a.lb}_R + \\
  & & r\sqrt{\Var_{\vt^*}[{\cal I}^b-{\cal I}^a]}+O(n^{-3/2}),\\[1ex]
  \Var_\vt[{\cal I}^b-{\cal I}^a] &=&
  \Var_\vt[{\cal I}^b] +
  \Var_\vt[{\cal I}^a] -
  2 \Cov_\vt[{\cal I}^b,{\cal I}^a],\\
  \Cov_\vt[{\cal I}^b,{\cal I}^a] &=&
  {1\over n+s}
  \sum_{ij\kdot}\u_{ij\kdot}\bigg(
    \log{\u^a_{ij}\over \u^a_{i\p}\u^a_{\p j}}
    \log{\u^b_{j\kdot}\over \u^b_{j\p}\u^b_{\p\kdot}}\bigg) \\ & &
  \hspace*{-8ex} -
  {1\over n+s}\bigg(\sum_{ij}\u^a_{ij}\log{\u^a_{ij}\over
    \u^a_{i\p}\u^a_{\p j}}\bigg)
    \bigg(\sum_{j\kdot}\u^b_{j\kdot}\log{\u^b_{j\kdot}\over
    \u^b_{j\p}\u^b_{\p\kdot}}\bigg) \;+\; O(n^{-2}).
\eqan
%These expressions with choice $t^*_{ij\dot{r}}=(d_\imath d_\jmath d_\kappa)^{-1}$ have
%been used in our experiments.
Variances are typically of order $1/n$, so for large $n$, credible
intervals $\up{\widetilde{\cal I}}-\underbartilde{\cal
I}=O(1/\sqrt{n})$ are much wider than expected intervals $\up
I-\low I=O(1/n)$.

%%%%%%%%%%%%%%%%%%%%%%%%%%%%%%%%%%%%%%%%%%%%%%%%%%%%%%%%%%%%%%%
\section{Conclusions}\label{secC}
%%%%%%%%%%%%%%%%%%%%%%%%%%%%%%%%%%%%%%%%%%%%%%%%%%%%%%%%%%%%%%%

This paper has tackled the problem to reliably infer trees from
data. We have provided an exact procedure that infers strong edges
in time $O(m^4)$, and have shown that it performs well in practice
on an example problem. We have also developed an approximate
algorithm that works in time $O(m^3)$.

Reliability follows from using the IDM, a robust inferential model
that rests on very weak prior assumptions. Working with the IDM
involves computing lower and upper estimates, i.e., solving global
optimization problems. These can hardly be tackled exactly, as
they are typically non-linear and non-convex. A substantial part
of the present work has been devoted to provide systematic
approximations to the exact intervals with a guaranteed worst case
of $O(\sigma^2)$. This was achieved by optimizing approximating
functions, obtained by Taylor-expanding the original objective
function. We have taken care to make these approximations
conservative, i.e., they always include the exact interval. This
is the necessary step to ultimately obtain over-cautious rather
than overconfident models.

More broadly speaking, the same approach has been used also for
another approximation, concerned with the representation level
chosen for the IDM. In principle, one might use the IDM for the
joint realization of all the $m$ random variables. In this paper
we have used one IDM for each bivariate (and tri-variate, in some
cases) realization. Using separate IDMs simplifies the treatment,
but it may give rise to global inconsistencies (in the same lines
of the discussion on comparing edges with a common vertex, in
Section~\ref{secCE}). However, their effect is only to make
$\mathcal{O}$ strictly include $\mathcal{O_T}$, thus producing an
excess of caution, as discussed in Section~\ref{secEDOSE}.

We have already reported two developments that follow naturally
from the work described above. The first involves the computation
of robust trees, which widens the scope of this paper to other
applications. The second is in the direction of even greater
robustness by providing robust credibile limits for mutual
information, which provide the user with a guarantee level on
the inferred dependency structures.

Other extensions of the present work could be considered that need
further research in order to be realized. Obviously, it would be worth
extending the work to the robust inference of more general
dependency structures. This could be achieved, for example, in a
way similar to Kleiter's work \cite{Kleiter99}. One could also
extend our approach to dependency measures other than mutual
information, like the statistical coefficient $\phi^2$
\cite[pp.~556--561]{KenStu67}. This would require new
approximations to be derived for the new index under the IDM, but
the first part of the paper on the detection of strong edges could
be applied as it is.

Another important extension could be realized by considering the
inference of dependency structures from incomplete samples. Recent
research has developed robust approaches to incomplete samples
that make very weak assumptions on the mechanism responsible for
the missing data \cite{Manski_book,RamSeb01b,zafjspi00b}. This
would be an important step towards realism and reliability in
structure inference.

\appendix
%%%%%%%%%%%%%%%%%%%%%%%%%%%%%%%%%%%%%%%%%%%%%%%%%%%%%%%%%%%%%%%
\section{\boldmath Properties of the digamma $\psi$ function}\label{secPsi}
%%%%%%%%%%%%%%%%%%%%%%%%%%%%%%%%%%%%%%%%%%%%%%%%%%%%%%%%%%%%%%%

The digamma function $\psi$ is defined as the logarithmic derivative of
the Gamma function. Integral representations for $\psi$ and its
derivatives are
\bqan
  \psi(z) &=& {d\ln\Gamma(z)\over dz}= {\Gamma'(z)\over\Gamma(z)} =
  \int_0^\infty\left[{e^{-t}\over t}-{e^{-zt}\over
  1-e^{-t}}\right] dt,
\\
  \psi^{(\ell)}(z) &=& (-1)^{\ell+1}\int_0^\infty{t^\ell e^{-zt}\over 1-e^{-t}} dt
  \qmbox{for} \ell>0.
\eqan
The $h$ function~(\ref{hex}) and its derivatives are
$
  h^{(\ell)}(\u) \;=\;
$
\beqn
  \u^{(\ell)}\psi(n\!+\!s\!+\!1) -
  \ell(n\!+\!s)^{\ell-1}\psi^{(\ell-1)}((n\!+\!s)\u\!+\!1) -
  \u(n\!+\!s)^\ell\psi^{(\ell)}((n\!+\!s)\u\!+\!1).
\eeqn
At argument $\u_i={n_i+st_i\over n+s}$ we get for $h$, $h'$ and $h''$
\bqan
  h(\u_i) &=& (n_i\!+\!st_i)[\psi(n\!+\!s\!+\!1)-\psi(n_i\!+\!st_i\!+\!1)]/(n\!+\!s),\\
  h'(\u_i) &=& \psi(n\!+\!s\!+\!1)-\psi(n_i\!+\!st_i\!+\!1)-(n_i\!+\!st_i)\psi'(n_i\!+\!st_i\!+\!1),\\
  h''(\u_i) &=& -2(n\!+\!s)\psi'(n_i\!+\!st_i\!+\!1)-(n_i\!+\!st_i)(n\!+\!s)\psi''(n_i\!+\!st_i\!+\!1),
\eqan
For integral arguments the following closed representations for
$\psi$, $\psi'$, and $\psi''$ exist:
\beqn
  \psi(n\!+\!1)=-\gamma+\sum_{i=1}^n{1\over i},\ifjournal\;\else\quad\fi
  \psi'(n\!+\!1)={\pi^2\over 6}-\sum_{i=1}^n{1\over i^2},\ifjournal\;\else\quad\fi
  \psi''(n\!+\!1)=-2\zeta(3)+2\sum_{i=1}^n{1\over i^3}
\eeqn
where $\gamma=0.5772156...$ is Euler's constant and
$\zeta(3)=1.202569...$ is Riemann's zeta function at 3. Closed
expressions for half-integer values and fast
approximations for arbitrary arguments also exist. The following
asymptotic expansion can be used if one is interested in $O(({s\over
n+s})^2)$ approximations only (and not rigorous bounds):
\beqn
  \psi(z+1)=\log z + {1\over 2z} - {1\over 12z^2} + O({1\over z^4}).
\eeqn
See \cite{Abramowitz:74} for details on the $\psi$ function and
its derivatives. From the above expressions one may show $h''<0$
and $h'''>0$.

%%%%%%%%%%%%%%%%%%%%%%%%%%%%%%%%%%%%%%%%%%%%%%%%%%%%%%%%%%%%%%%
%         Bibliography                                        %
%%%%%%%%%%%%%%%%%%%%%%%%%%%%%%%%%%%%%%%%%%%%%%%%%%%%%%%%%%%%%%%

\begin{small}

\end{small}

\ifjournal\end{article}\fi

\begin{thebibliography}{GCSR95}

\bibitem[AS74]{Abramowitz:74}
M.~Abramowitz and I.~A. Stegun, editors.
\newblock {\em Handbook of Mathematical Functions}.
\newblock Dover publications, inc., 1974.

\bibitem[AVH04]{aron2004}
I.~D. Aron and P.~{V}an {H}entenryck.
\newblock On the complexity of the robust spanning tree problem with interval
  data.
\newblock {\em Operations Research Letters}, 32:36--40, 2004.

\bibitem[Ber01]{Ber01}
J.-M. Bernard.
\newblock Non-parametric inference about an unknown mean using the imprecise
  {D}irichlet model.
\newblock In G.~{d}e Cooman, T.~Fine, and T.~Seidenfeld, editors, {\em
  {ISIPTA}'01}, pages 40--50, The Netherlands, 2001. Shaker Publishing.

\bibitem[Ber05]{bernard2005}
J.-M. Bernard.
\newblock An introduction to the imprecise {D}irichlet model for multinomial
  data.
\newblock {\em International Journal of Approximate Reasoning},
  39(2--3):123--150, 2005.

\bibitem[CL68]{ChowLiu68}
C.~K. Chow and C.~N. Liu.
\newblock Approximating discrete probability distributions with dependence
  trees.
\newblock {\em {IEEE} Transactions on Information Theory}, IT-14(3):462--467,
  1968.

\bibitem[FGG97]{FrGeGo97}
N.~Friedman, D.~Geiger, and M.~Goldszmidt.
\newblock {B}ayesian networks classifiers.
\newblock {\em Machine Learning}, 29(2/3):131--163, 1997.

\bibitem[GCSR95]{Gelman:95}
A.~Gelman, J.~B. Carlin, H.~S. Stern, and D.~B. Rubin.
\newblock {\em Bayesian Data Analysis.}
\newblock Chapman, 1995.

\bibitem[Hal48]{Haldane:48}
J.~B.~S. Haldane.
\newblock The precision of observed values of small frequencies.
\newblock {\em Biometrika}, 35:297--300, 1948.

\bibitem[Hut01]{Hut01}
M.~Hutter.
\newblock Distribution of mutual information.
\newblock In T.~G. Dietterich, S.~Becker, and Z.~Ghahramani, editors, {\em
  Proceedings of NIPS*2001}, Cambridge, MA, 2001. MIT Press.

\bibitem[Hut03]{Hut03}
M.~Hutter.
\newblock Robust estimators under the {I}mprecise {D}irichlet {M}odel.
\newblock In {\em Proc. 3rd International Symposium on Imprecise Probabilities
  and Their Application ({ISIPTA-2003})}, volume~18 of {\em Proceedings in
  Informatics}, pages 274--289, Canada, 2003. Carleton Scientific.

\bibitem[HZ05]{HutZaf05}
M.~Hutter and M.~Zaffalon.
\newblock Distribution of mutual information from complete and incomplete data.
\newblock {\em Computational Statistics \& Data Analysis}, 48(3):633--657,
  2005.

\bibitem[Jef46]{Jeffreys:46}
H.~Jeffreys.
\newblock An invariant form for the prior probability in estimation problems.
\newblock In {\em Proc. Royal Soc. London (A)}, volume 186, pages 453--461,
  1946.

\bibitem[KJ56]{Kruskal56}
J.~B. Kruskal~Jr.
\newblock On the shortest spanning subtree of a graph and the traveling
  salesman problem.
\newblock In {\em Proc. Am. Math. Soc.}, volume~7, pages 48--50, 1956.

\bibitem[KL51]{KulLei51}
S.~Kullback and R.~A. Leiber.
\newblock On information and sufficiency.
\newblock {\em Ann. Math. Statistics}, 22:79--86, 1951.

\bibitem[Kle99]{Kleiter99}
G.~D. Kleiter.
\newblock The posterior probability of {B}ayes nets with strong dependences.
\newblock {\em Soft Computing}, 3:162--173, 1999.

\bibitem[KS67]{KenStu67}
M.~G. Kendall and A.~Stuart.
\newblock {\em The Advanced Theory of Statistics}.
\newblock Griffin, London, 1967.
\newblock 2nd edition.

\bibitem[Kul68]{Kul68}
S.~Kullback.
\newblock {\em Information Theory and Statistics}.
\newblock Dover, 1968.

\bibitem[Man02]{Manski_book}
C.~Manski.
\newblock {\em Partial Identification of Probability Distributions}.
\newblock Draft book, Department of Economics, Northwestern University, USA,
  2002.

\bibitem[Mon0X]{montemanni2005a}
R.~Montemanni.
\newblock A {B}enders decomposition approach for the robust spanning tree
  problem with interval data.
\newblock {\em European Journal of Operational Research}, 200X.
\newblock Forthcoming.

\bibitem[Pea88]{Pearl88}
J.~Pearl.
\newblock {\em Probabilistic Reasoning in Intelligent Systems: Networks of
  Plausible Inference}.
\newblock Morgan Kaufmann, San Mateo, 1988.

\bibitem[Per47]{Perks:47}
W.~Perks.
\newblock Some observations on inverse probability.
\newblock {\em J. Inst. Actuar.}, 73:285--312, 1947.

\bibitem[PS82]{PapSte82}
H.~Papadimitriou and K.~Steiglitz.
\newblock {\em Combinatorial Optimization: Algorithms and Complexity}.
\newblock Prentice Hall, New York, 1982.

\bibitem[RS01]{RamSeb01b}
M.~Ramoni and P.~Sebastiani.
\newblock Robust learning with missing data.
\newblock {\em Machine Learning}, 45(2):147--170, 2001.

\bibitem[VP90]{VerPea90}
T.~Verma and J.~Pearl.
\newblock Equivalence and synthesis of causal models.
\newblock In P.~P. Bonissone, M.~Henrion, L.~N. Kanal, and J.~F. Lemmer,
  editors, {\em UAI'90}, pages 220--227, New York, 1990. Elsevier.

\bibitem[Wal91]{Wal91}
P.~Walley.
\newblock {\em Statistical Reasoning with Imprecise Probabilities}.
\newblock Chapman and Hall, New York, 1991.

\bibitem[Wal96]{Wal96b}
P.~Walley.
\newblock Inferences from multinomial data: learning about a bag of marbles.
\newblock {\em J. R. Statist. Soc. B}, 58(1):3--57, 1996.

\bibitem[WW95]{Wolpert:95}
D.~H. Wolpert and D.~R. Wolf.
\newblock Estimating functions of distributions from a finite set of samples.
\newblock {\em Physical Review E}, 52(6):6841--6854, 1995.

\bibitem[YKP01]{YamKarPin01}
H.~Yaman, O.~E. Kara\c{s}an, and M.~\c{C}. Pinar.
\newblock The robust spanning tree problem with interval data.
\newblock {\em Operations Research Letters}, 29:31--40, 2001.

\bibitem[Zaf02]{zafjspi00b}
M.~Zaffalon.
\newblock Exact credal treatment of missing data.
\newblock {\em Journal of Statistical Planning and Inference}, 105(1):105--122,
  2002.

\end{thebibliography}
\end{document}